\documentclass[10pt,twocolumn,letterpaper]{article}

\usepackage[pagenumbers]{cvpr} 

\usepackage{graphicx}
\usepackage{amsmath}
\usepackage{amssymb}
\usepackage{booktabs}
\usepackage{multirow}
\usepackage{color, colortbl}
\usepackage{float}
\usepackage{arydshln}
\usepackage{subcaption}
\usepackage{comment}
\usepackage{arydshln}
\usepackage{flushend}
\usepackage{adjustbox}
\usepackage[dvipsnames]{xcolor}

\newcommand*\circled[1]{\raisebox{.5pt}{\textcircled{\raisebox{-.9pt} {\small #1}}}}

\usepackage[pagebackref,breaklinks,colorlinks]{hyperref}

\usepackage[capitalize]{cleveref}
\crefname{section}{Sec.}{Secs.}
\Crefname{section}{Section}{Sections}
\Crefname{table}{Table}{Tables}
\crefname{table}{Tab.}{Tabs.}

\definecolor{Gray}{gray}{0.9}

\def\cvprPaperID{10350} 
\def\confName{CVPR}
\def\confYear{2023}

\begin{document}

\title{Zero Shot Object Detection}
\title{Frustratingly Simple but Effective Zero-shot Detection and Segmentation: \\ Analysis and a Strong Baseline}

\author{
Siddhesh Khandelwal$^{1,2}$ \hspace{0.25in} Anirudth Nambirajan$^{4}$ \hspace{0.25in} Behjat Siddiquie$^{4}$\\
Jayan Eledath$^{4}$ \hspace{0.25in} Leonid Sigal$^{1,2,3}$\\
$^1$University of British Columbia \hspace{0.25in} $^2$Vector Institute for AI \hspace{0.25in} $^3$CIFAR AI Chair \hspace{0.25in} $^4$Amazon  \\
\texttt{\{skhandel, lsigal\}@cs.ubc.ca} \hspace{0.25in}
\texttt{\{anirudtn, behjats, eledathj\}@amazon.com} \\
}


\twocolumn[{
\renewcommand\twocolumn[1][]{#1}
\vspace*{-12mm}
\maketitle

\begin{center}
    \centering
    \includegraphics[width=0.9\textwidth]{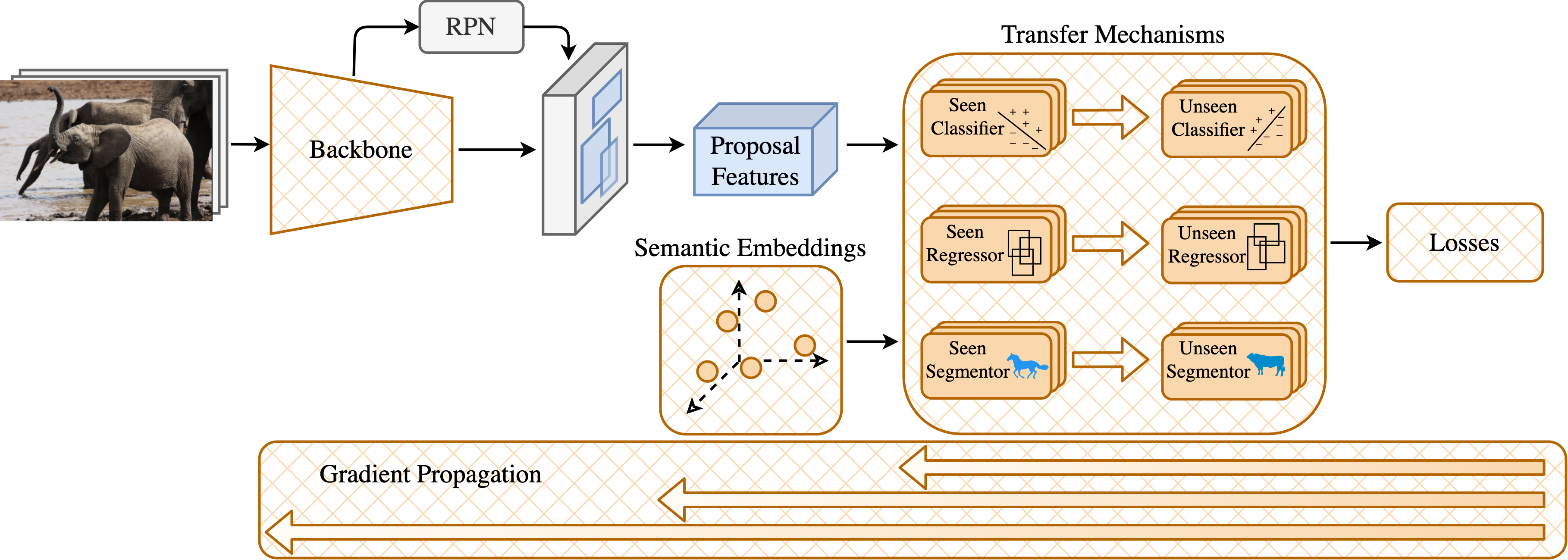}
    \captionof{figure}{{\bf Design Choices for Zero-Shot Detection and Segmentation.} Illustrated is an abstracted generalized architecture for zero-shot detection and segmentation. Each component in \textcolor{orange}{orange} denotes a possible design choice for which multiple alternatives exist. By carefully considering and ablating these design choices we construct a simple yet highly effective architecture that achieves state-of-the-art performance. We propose this model as a new {\em strong} baseline for the zero-shot detection and segmentation tasks.
    }
    \label{fig:teaser}
\end{center}
}]

\begin{abstract}
   Methods for object detection and segmentation often require abundant instance-level annotations for training, which are time-consuming and expensive to collect. To address this, the task of zero-shot object detection (or segmentation) aims at learning effective methods for identifying and localizing object instances for the categories that have no supervision available. Constructing architectures for these tasks requires choosing from a myriad of design options, ranging from the
   form of the class encoding 
   used to transfer information from seen to unseen categories, to the nature of the function being optimized for learning. In this work, we extensively study these design choices, and carefully construct a simple yet extremely effective zero-shot recognition method. Through extensive experiments on the MSCOCO \cite{lin2014microsoft} dataset on object detection and segmentation, we highlight that our proposed method outperforms existing, considerably more complex, architectures.
   Our findings and method, which we propose as a competitive future baseline, point towards the need to revisit some of the recent design trends in zero-shot detection / segmentation. 
\end{abstract}

\section{Introduction}
\vspace{-0.5em}
\label{sec:intro}
Advancements in CNN based deep learning architectures over the years have lead to significant improvements in computer vision recognition tasks such as object detection \cite{liu2016ssd,redmon2017yolo9000,ren2015faster} and segmentation \cite{chen2018masklab,he2017mask}, both in terms of recognition quality and speed. However, traditional CNN-based approaches often rely on the availability of abundant supervision for learning, which are both time-consuming and expensive to gather \cite{hoffman2014lsda, laradji2019masks}. This effect is more pronounced for instance-level annotations like bounding boxes and segmentation masks \cite{Bearman_2019_ECCV}, 
thus making scaling of object detection and segmentation to new categories challenging.

As a consequence, research into learning methods that generalize to categories with no available supervision -- referred to as \emph{zero-shot learning} -- has gained significant traction, wherein the focus is to develop techniques for information transfer from the supervision abundant \emph{seen} categories to the supervision-absent \emph{unseen} categories. Although there has been considerable work towards zero-shot learning on image-level tasks such as image classification \cite{changpinyo2016synthesized,kodirov2017semantic,rahman2018unified,xian2017zero,zhang2015zero,zhang2016zero,zablocki2019context}, more granular recognition problems such as zero-shot detection and segmentation are relatively unexplored \cite{huang2022robust,zheng2021zero,zheng2020background,hayat2020synthesizing,rahman2018polarity}. These instance-level tasks are naturally more challenging as, in addition to accurately identifying the objects present in an image, methods are required to precisely localize them -- either via bounding boxes or segmentation masks. 

Constructing architectures for the zero-shot detection (or segmentation) requires making certain critical design choices that directly impact performance. These include decisions on the -- \circled{i} model characteristics like capacity, and mechanisms for information transfer from seen to unseen categories, \circled{ii} learning dynamics like the choice of loss function, and \circled{iii} inference procedure like selecting the appropriate trade-off between seen and unseen category performance. Existing approaches often explore only one of these critical options, leading to sub-optimal selection for the remaining choices. Concretely, the recent focus of zero-shot detection (or segmentation) has been towards creating complex models, both in terms of the number of parameters \cite{huang2022robust,hayat2020synthesizing} and the use of intricate modules aimed at better information transfer from seen to unseen categories \cite{huang2022robust,hayat2020synthesizing,zheng2020background,zheng2021zero,gupta2020multi}. 

In this work, we argue that this complexity is unnecessary, and propose a simple solution to zero-shot detection (and instance segmentation) by extensively exploring the aforementioned design decisions. More specifically, our model specifications are dictated by a set of carefully constructed ablation studies, one for each possible choice. Our approach adopts a two-step training scheme, wherein the first step involves training an object detector (or segmentor) like Faster R-CNN \cite{ren2015faster} (or Masked R-CNN \cite{he2017mask}) on the \emph{seen} categories with abundant instance-level annotations. The second stage \emph{fine-tunes} a projection layer, trained on the \emph{seen} categories to learn a transformation from image features to a semantically meaningful space. Information transfer from the \emph{seen} to \emph{unseen} categories, for the classifiers, detectors, and segmentors, is achieved by leveraging \emph{normalized} category-name semantic embeddings obtained from unsupervised approaches like GloVe \cite{pennington2014glove} or ConceptNet \cite{speer2017conceptnet}. This straightforward fine-tuning approach, when trained using the cross-entropy loss function, outperforms most existing methods that are significantly more complex in design. 

The ablations and positive performance of our proposed simple approach motivate the  need to more broadly revisit the research direction in the field of zero-shot detection (and segmentation). For example, although a large amount of existing work focuses on improving performance through complex architectures \cite{huang2022robust,hayat2020synthesizing}, we find that the choice of semantic embeddings (like GloVe \cite{pennington2014glove}) has the largest impact on performance. Despite its untapped potential, this direction is seldom explored in the literature.

\noindent
\textbf{Contributions. }Our foremost contribution is a simple yet extremely effective architecture for zero shot detection and segmentation. The characteristics for our proposed method are carefully curated via extensive exploration over various critical design decisions, and is trained with a straightforward two-step process. We demonstrate the efficacy of our approach by thorough evaluations on the MSCOCO \cite{lin2014microsoft} dataset, which show that our proposed solution outperforms existing, considerably more complex, architectures. On the basis of these results, we argue for the need to revisit some of the recent design trends in the field of zero shot detection and segmentation, wherein our proposed method serves as a competitive baseline.

\vspace{-0.5em}
\section{Related Work}
\vspace{-0.5em}

\begin{figure*}[t]
    \centering
    \includegraphics[width=0.9\textwidth]{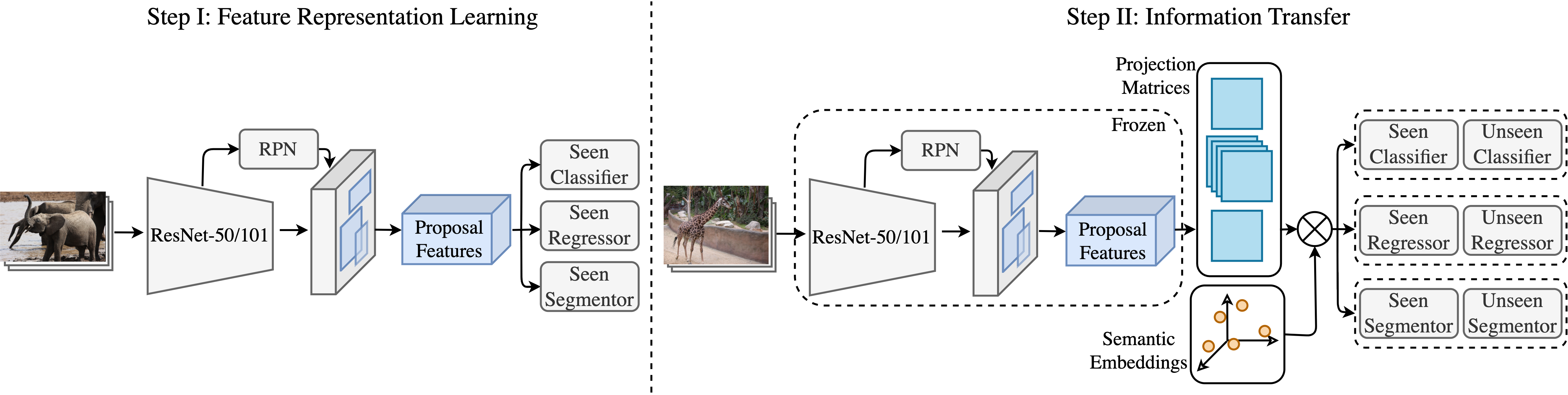}
    \vspace{-0.15in}
    \caption{{\bf Proposed Approach with Two-step Training.} The first step trains the learnable parameters of a Faster \cite{ren2015faster} / Mask \cite{he2017mask} RCNN architecture using the seen category annotations. The second step freezes these parameters, and learns the projection matrices which, along with the semantic embeddings, generates embedding-aware classifiers, regressors, and segmentors for both seen and unseen categories.}
    \label{fig:twostage}
    \vspace{-0.15in}
\end{figure*}

\vspace{0.2em}
\noindent
\textbf{Zero-shot Detection. } Introduced in \cite{bansal2018zero, demirel2018zero}, the task of zero-shot detection (ZSD) poses the challenge of localizing unseen objects within an image. The majority of existing work in this field has been towards modifying model construction for improved performance \cite{gupta2020multi,hayat2020synthesizing,huang2022robust,zheng2020background,zheng2021zero,zhu2020don}. Gupta \etal \cite{gupta2020multi} learn a multi-head network to disentangle visual and semantics spaces, each separately identifying object categories, which are subsequently ensembled. \cite{bansal2018zero,zheng2020background,zheng2021zero} propose methods to better learn semantic embeddings for the non-object (or background) category in an attempt to better distinguish them from the unseen categories. Bansal \etal \cite{bansal2018zero} employ an iterative expectation maximization (EM) like procedure to generate a background embedding vector. \cite{zheng2020background,zheng2021zero} both modify the region proposal network (RPN) to learn embeddings to accurately perform the foreground-background binary classification task. Authors in \cite{hayat2020synthesizing,huang2022robust,zhu2020don} utilize generative model-based methods, wherein the aim is to synthesize features for unseen categories which can be used downstream to learn better classifiers.
The focus of \cite{rahman2018zero,rahman2018polarity,rahman2019transductive} is improving learning procedures for ZSD. Experimenting with loss functions, \cite{rahman2018zero} propose a max-margin and clustering based loss, whereas \cite{rahman2018polarity} suggest the use of a novel polarity loss to encourage better visual-semantic alignment. Rahman \etal \cite{rahman2019transductive}, on the other hand, study the transductive learning learning paradigm for ZSD, and pseudo-label unseen category images to generate additional training data. Note that ZSD, which is the setting used in this work, assumes \emph{no} unseen category supervision. On the other hand, works in the open vocabulary detection literature \cite{gu2021open,zareian2021open,du2022learning} assume unseen category information through pretrained models, or visuo-lingual annotations, and are therefore not directly comparable.

\vspace{0.2em}
\noindent
\textbf{Zero-shot Segmentation. } Zero-shot segmentation is a relatively unexplored sub-field \cite{bucher2019zero, ding2022decoupling,hu2020uncertainty,hu2018learning,huynh2022open,Kato_2019_ICCV,khandelwal2021unit,zhao2017open,zheng2021zero}. Existing work has largely focused on  zero-shot semantic segmentation \cite{bucher2019zero,ding2022decoupling,hu2020uncertainty,hu2018learning,Kato_2019_ICCV,zhao2017open}, wherein the aim is to accurately label each pixel in an image. Zhao \etal \cite{zhao2017open} utilize WordNet \cite{miller1995wordnet} hypernym/hyponym relations to segment images. \cite{bucher2019zero,Kato_2019_ICCV,ding2022decoupling} focus on changing model construction to improve the information transfer from seen to unseen categories. Bucher \etal \cite{bucher2019zero} leverage synthetic features from a generative model to train a classifier for unseen categories, whereas Kato \etal \cite{Kato_2019_ICCV} learn a variational mapping over category-name embeddings from the semantic to visual space. Ding \etal \cite{ding2022decoupling} instead decompose the zero-shot semantic segmentation problem into sub-tasks of class-agnostic grouping of pixels and subsequent classification over these groupings. Hu \etal \cite{hu2020uncertainty} emphasize improving learning dynamics by proposing uncertainty aware losses to mitigate the adverse effect of noisy seen category information on model performance. Works in \cite{huynh2022open,khandelwal2021unit,zheng2021zero} study the task of zero-shot instance segmentation, where the goal is to identify and generate accurate masks for individual object instances. \cite{huynh2022open,khandelwal2021unit} both assume the availability of image-level annotations for unseen categories, either in the form of captions or labels. The work in \cite{zheng2021zero}, that learns a separate background category embedding, makes no such assumption, and resembles the setting used in this work.

\vspace{0.2em}
\noindent
\textbf{Semantic Embeddings. }Most approaches in ZSL use class-label semantic embeddings as the building block for efficient information transfer from seen to unseen categories \cite{gupta2020multi,hayat2020synthesizing,huang2022robust,zheng2020background,zheng2021zero,zhu2020don,bucher2019zero, ding2022decoupling,hu2020uncertainty,hu2018learning,huynh2022open,Kato_2019_ICCV,khandelwal2021unit,zhao2017open}. Methods to generate these embeddings have been widely explored in NLP and computer vision literature \cite{pennington2014glove,mikolov2013efficient,speer2017conceptnet,radford2021learning,reimers2019sentence,devlin2018bert}. These embeddings effectively capture semantic and syntactic similarities between words (or sentences). Therefore, in this work, we also utilize these embeddings to project images and category labels into a common feature space, thus allowing detection (or segmentation) of unseen categories.

\vspace{-0.5em}
\section{Problem Formulation}
\vspace{-0.5em}
\label{sec:formulation}
In this section we formally introduce the zero-shot detection (ZSD) / instance segmentation (ZSI) setup. The tasks assume two sets of disjoint categories - namely \emph{seen} $\mathcal{C}^s$ and \emph{unseen} $\mathcal{C}^u$, where $\mathcal{C}^s \cap \mathcal{C}^u = \varnothing$. For the seen categories, consistent with existing work, we assume the availability of abundant instance-level annotations $\mathcal{D}^s = \{(\mathbf{x}_i, \mathbf{c}_i, \mathbf{y}_i)\}$, where $\mathbf{x}_i$ is an input image, $\mathbf{c}_i = \{c_{i,j}\}$ are seen category labels, and $\mathbf{y}_i = \{\textbf{bbox}_{i,j}\}$ or $\mathbf{y}_i = \{\textbf{mask}_{i,j}\}$ are the corresponding bounding boxes and/or masks for each instance $j$ in image $i$. Note that, for the unseen categories, no supervision is provided, \ie $\mathcal{D}^u = \{\mathbf{x}_i\}$. However, it is assumed that semantic embeddings $\mathbf{E}^s \in \mathbb{R}^{|\mathcal{C}^s| \times d}$ and $\mathbf{E}^u \in \mathbb{R}^{|\mathcal{C}^u| \times d}$ for the seen and unseen categories respectively are known, where $d$ is the embedding dimensionality. We define $\mathbf{E} = \{\mathbf{E}^s,\mathbf{E}^u\}$ as the set of all available embeddings. At inference, the goal is to generate category-labels, bounding boxes, and possibly segmentation masks for the unseen categories. Depending on the scenario, the test set may contain only unseen objects (ZSD / ZSI setup), or both seen and unseen objects (GZSD\footnote{The G stands for {\em generalized}.} / GZSI setup).

\vspace{-0.5em}
\section {Approach}
\vspace{-0.5em}
\label{sec:approach}
Constructing an effective architecture for the task of zero-shot detection (or segmentation) necessitates making certain critical design decisions. By extensively exploring these choices and carefully selecting model components, we propose a simple solution for ZSD and ZSI. 

For the task of zero-shot detection, our proposed approach, illustrated in Figure \ref{fig:twostage}, adopts the popular two-stage detector Faster-RCNN \cite{ren2015faster}. The Faster-RCNN architecture consists of four learnable components, namely -- \circled{i} the backbone (like ResNet \cite{he2016deep} or VGG \cite{simonyan2014very}), \circled{ii} the region proposal network (RPN), \circled{iii} proposal-level feature extractor, and \circled{iv} classifier and class-aware regressor heads. The first stage of the aforementioned two-stage detector uses the backbone to extract features $\mathbf{\bar{x}}_i$ from an input image $\mathbf{x_i}$, which are subsequently utilized by the RPN to generate class-agnostic object region proposals $\{\mathbf{pbox}_{i,j}\}$. The second stage involves a detection pipeline wherein the proposal-level feature extractor $f_{\mathbf{W}^{prop}}$ performs region-of-interest (RoI) pooling, and generates proposal features $\mathbf{z}_{i,j} = f_{\mathbf{W}^{prop}}\left(\texttt{RoIPool}\left(\mathbf{\bar{x}}_i,\mathbf{pbox}_{i,j} \right)\right)$ for each proposal $j$. The classifier head learns to label the proposal feature $\mathbf{z}_{i,j}$ into one of the seen categories, and the class-aware regressor head utilizes the proposal feature $\mathbf{z}_{i,j}$ to refine the bounding box proposals $\{\mathbf{pbox}_{i,j}\}$. Note that, for zero-shot instance segmentation, our method builds on the Mask-RCNN \cite{he2017mask} model that additionally learns a segmentation head to generate masks for the detected objects given $\mathbf{z}_{i,j}$. Our approach disentangles the learning of feature representations and transfer of information from seen to unseen categories, and is therefore trained in two steps.

\vspace{0.2em}
\noindent
\textbf{Feature Representation Learning. } The first step entails training the aforementioned learnable components of the Faster/Mask RCNN architecture on seen category instance-level data $\mathcal{D}^s$, guided by the losses described in \cite{ren2015faster,he2017mask}. 

\vspace{0.2em}
\noindent
\textbf{Information Transfer from Seen to Unseen Categories. }In the second step, we obtain the classification, regression, and segmentation heads for the unseen categories by leveraging the relationships between the seen and unseen semantic embeddings $\mathbf{E}^s$ and $\mathbf{E}^u$. This is achieved via learning a joint visual-semantic feature space that facilitates interactions between image features and class embeddings. 

\vspace{0.2em}
\noindent
\textbf{Classifier. }The embedding-aware classifier $f_{\mathbf{W}^{cls}_{seen}}$ for the seen categories uses a projection matrix $\mathbf{W}^{cls}$ that projects the proposal features $\mathbf{z}_{i,j}$ to the embedding feature space,
{
\setlength{\abovedisplayskip}{4pt}
\setlength{\belowdisplayskip}{4pt}
\setlength{\abovedisplayshortskip}{4pt}
\setlength{\belowdisplayshortskip}{4pt}
\begin{align}
    f_{\mathbf{W}^{cls}_{seen}}(\mathbf{{z}}_{i,j}) = \left(\mathbf{W}^{cls}\mathbf{{z}}_{i,j}\right)\left( \frac{\mathbf{\overline{E}}^s}{\lVert\mathbf{\overline{E}}^s\rVert_2}\right)^T
\end{align}
where $\lVert . \rVert_2$ is the L2 norm. $\mathbf{\overline{E}}^s = [\mathbf{E}^s, \mathbf{b}] \in \mathbb{R}^{(|\mathcal{C}^s| + 1)\times d}$ is constructed by augmenting the background embedding $\mathbf{b} \in \mathbb{R}^d$ that is learned alongside the projection matrices. 
We leverage the projection matrix $\mathbf{W}^{cls}$ to simply define the embedding-aware classifier 
\begin{align}
    f_{\mathbf{W}^{cls}_{unseen}}(\mathbf{{z}}_{i,j}) = \left(\mathbf{W}^{cls}\mathbf{{z}}_{i,j}\right)\left( \frac{\mathbf{{E}}^u}{\lVert\mathbf{{E}}^u\rVert_2}\right)^T
\end{align}
During inference, the probabilities $\mathbf{p}_{i,j}$ over all categories is expressed as,
\begin{align}
    \label{eq:classifier-prob}
    \mathbf{p}_{i,j} = \sigma \left( \left[f_{\mathbf{W}^{cls}_{seen}}(\mathbf{{z}}_{i,j}), f_{\mathbf{W}^{cls}_{unseen}}(\mathbf{{z}}_{i,j})  \right] \right)
\end{align}
where $\sigma$ is the softmax function, and $[.,.]$ represents the concatenation operation. 
}

\vspace{0.2em}
\noindent
\textbf{Regressor. }A valid bounding box can be defined by the top-left and bottom-right $(x,y)$ coordinates. The embedding-aware regressor $f_{\mathbf{W}^{reg}_{seen}}$ for the seen categories therefore uses four projection matrices $\mathbf{W}^{reg}_r, r \in [1,4]$, where a pair of matrices generate one of the two required coordinates,  
{
\setlength{\abovedisplayskip}{4pt}
\setlength{\belowdisplayskip}{4pt}
\setlength{\abovedisplayshortskip}{4pt}
\setlength{\belowdisplayshortskip}{4pt}
\begin{align}
    f_{\mathbf{W}^{reg}_{seen}}(\mathbf{{z}}_{i,j}) = \left\{\left(\mathbf{W}^{reg}_r\mathbf{{z}}_{i,j}\right)\left( \frac{\mathbf{{E}}^s}{\lVert\mathbf{{E}}^s\rVert_2}\right)^T\right\};\; r\in [1, 4].
\end{align}
The regressor for the unseen categories $f_{\mathbf{W}^{reg}_{unseen}}$ is analogously defined as,
\begin{align}
    f_{\mathbf{W}^{reg}_{unseen}}(\mathbf{{z}}_{i,j}) = \left\{\left(\mathbf{W}^{reg}_r\mathbf{{z}}_{i,j}\right)\left( \frac{\mathbf{{E}}^u}{\lVert\mathbf{{E}}^u\rVert_2}\right)^T\right\};\; r\in [1, 4].
\end{align}
}

\vspace{0.2em}
\noindent
\textbf{Segmentor.} The segmentation head within the Mask-RCNN \cite{he2017mask} architecture employs a separate proposal-level feature extractor $f_{\mathbf{W}^{mask}}$ to extract relevant spatial features $\mathbf{z}^m_{i,j} \in \mathbb{R}^{n \times n \times t}$ for segmentation mask generation, where $\mathbf{z}^m_{i,j} = f_{\mathbf{W}^{mask}}\left(\texttt{RoIPool}\left(\mathbf{\bar{x}}_i, \mathbf{pbox}_{i,j}\right)\right)$. Here $n$ and $t$ represent the spatial and feature dimension respectively. For a particular spatial coordinate $(x,y)$ feature $\mathbf{{z}}^m_{i,j}[x,y]$, the embedding-aware segmentor $f_{\mathbf{W}^{seg}_{seen}}$ for the seen categories is implemented as follows,
{
\setlength{\abovedisplayskip}{4pt}
\setlength{\belowdisplayskip}{4pt}
\setlength{\abovedisplayshortskip}{4pt}
\setlength{\belowdisplayshortskip}{4pt}
\begin{align}
    f_{\mathbf{W}^{seg}_{seen}}(\mathbf{{z}}^m_{i,j}[x,y]) = \left\{\left(\mathbf{W}^{seg} \mathbf{{z}}^m_{i,j}[x,y]\right)\left( \frac{\mathbf{{E}}^s}{\lVert\mathbf{{E}}^s\rVert_2}\right)^T\right\}
\end{align}
The segmentor for the unseen categories $f_{\mathbf{W}^{seg}_{unseen}}$ for the spatial coordinate $(x,y)$ follows a similar formulation,
\begin{align}
    f_{\mathbf{W}^{seg}_{unseen}}(\mathbf{{z}}^m_{i,j}[x,y]) = \left\{\left(\mathbf{W}^{seg} \mathbf{{z}}^m_{i,j}[x,y]\right)\left( \frac{\mathbf{{E}}^u}{\lVert\mathbf{{E}}^u\rVert_2}\right)^T\right\}
\end{align}
}

The training in the second step involves \emph{freezing} all the learnable parameters for the Faster/Mask RCNN architecture, and only learning the matrices ($\mathbf{W}^{cls}$, $\mathbf{W}_{r}^{reg}$, $\mathbf{W}^{seg}$). For the classifier this is done via a cross-entropy loss, the detector utilizes a smooth-L1 loss, and the segmentation head uses a pixel-level binary cross-entropy loss.

During inference, we utilize the aforementioned classifiers, regressors, and segmentors to generate predictions for proposals obtained from the RPN. A category-wise non-maximum suppression (NMS) \cite{ren2015faster} is applied over these predictions to remove overlapping outputs. We additionally define a threshold $\beta$ to allow the model to flexibly bias itself towards the unseen categories without the need for re-training. Specifically, for the seen categories, we remove any prediction with a classifier confidence less that $\beta$, \ie $\mathbf{p}_{i,j}^s < \beta; s \in \mathcal{C}^s$. $\mathbf{p}_{i,j}^s$ refers to the probability for the seen category $s$ (Equation \ref{eq:classifier-prob}). Therefore, as we generate a fixed number of predictions (\eg 100), increasing $\beta$ biases the model towards unseen categories. Such thresholding has been empirically used in literature \cite{rahman2018polarity,zheng2020background}, and serves as a measure to counter-balance the inherent bias against unseen categories due to the lack of training examples.

In the next section we justify our model architecture by exploring certain critical choices through ablations.

\subsection{Design Decisions}
\label{sec:designchoices}
\vspace{-0.5em}
The design decisions to construct the aforementioned model can be segregated into three groups, depending on whether they impact model characteristics, learning dynamics, or the inference procedure. 

\vspace{0.2em}
\noindent
\textbf{Model Characteristics. } We explore the impact of -- \circled{i} the capacity of the backbone, \circled{ii} the source of category embeddings $\mathbf{E}$, \circled{iii} the information transfer mechanism used to obtain the unseen category regressor and segmentor, and \circled{iv} the formulation of the background semantic embeddings.

\vspace{0.2em}
\noindent
\textbf{Learning Dynamics. } We analyze the effects of -- \circled{i} the type of loss used to train the classifier, and \circled{ii} fine-tuning the learnable parameters within the Faster-RCNN \cite{ren2015faster} (or Mask-RCNN \cite{he2017mask}) framework.

\vspace{0.2em}
\noindent
\textbf{Inference Procedure. }We study the seen-unseen category performance trade-off. This trade-off can be easily achieved by varying the parameter $\beta$ during inference. This affords the model the flexibility to operate on a spectrum of performance values without requiring any re-training.

To better understand the impact of a particular design decision, we ablate our proposed model by changing only the corresponding element, while keeping other components the same. Unless otherwise specified, the ablations are done using the Faster-RCNN \cite{ren2015faster} architecture with a ResNet-50 \cite{he2016deep} backbone, the Word2Vec \cite{mikolov2013efficient} semantic embeddings, and a threshold parameter $\beta=0.05$. The results are shown on the MSCOCO dataset \cite{lin2014microsoft} using the seen-unseen split proposed in \cite{bansal2018zero}, with $48$ seen and $17$ unseen categories. Further details on the dataset are provided in Section \ref{sec:experiment}. We generate $100$ predictions per image from the model, and report the mean average precision (mAP), and $\text{Recall}@100$ measured at $\text{IoU}{=}0.5$, and the harmonic mean (HM) over the seen and unseen category performance. The results are reported on both the ZSD and GZSD setups (see Section \ref{sec:formulation}). Note that, due to the similarities between the Faster-RCNN \cite{ren2015faster} and Mask-RCNN \cite{he2017mask} architectures, the observations from these ablations are directly applicable to the ZSI/GZSI tasks as well.  

\vspace{-1em}
\subsubsection{Backbone Capacity}
\vspace{-0.5em}
The capacity of the backbone can be varied by changing the depth of the ResNet \cite{he2016deep} architecture. The choice of backbone is often constraint to ResNet-50 \cite{zheng2020background, rahman2018polarity, rahman2019transductive} or ResNet-101 \cite{hayat2020synthesizing, huang2022robust, yan2022semantics} in existing literature. The table below analyzes the benefits of a deeper backbone architecture.

\vspace{-1em}
\begin{table}[H]
\setlength{\aboverulesep}{0pt}
\setlength{\belowrulesep}{0pt}
\setlength{\extrarowheight}{.75ex}
\setlength{\tabcolsep}{4.2pt}
\centering
\begin{tabular}{c|c|c|ccc}
\toprule
\toprule
\rowcolor{Gray}
 &  &  & \multicolumn{3}{c}{GZSD}\\ \cline{4-6} 
\rowcolor{Gray}
\multirow{-2}{*}{Metric} & \multirow{-2}{*}{Backbone} & \multirow{-2}{*}{ZSD} & \multicolumn{1}{c|}{Seen} & \multicolumn{1}{c|}{Unseen} & HM \\ \hline
\multirow{2}{*}{mAP}    & ResNet-50 \cite{he2016deep}                          &  $13.9$                    & \multicolumn{1}{c|}{$47.3$}     & \multicolumn{1}{c|}{$9.4$}       &  $15.7$  \\
                        & ResNet-101 \cite{he2016deep}                           &   $\mathbf{14.8}$                  & \multicolumn{1}{c|}{$\mathbf{48.9}$}     & \multicolumn{1}{c|}{$\mathbf{10.2}$}       & $\mathbf{16.9}$   \\ \cdashline{1-6}
\multirow{2}{*}{Recall} &   ResNet-50 \cite{he2016deep}                        &    $59.7$                  & \multicolumn{1}{c|}{$68.5$}     & \multicolumn{1}{c|}{$55.1$}       & $61.1$   \\
                        &  ResNet-101 \cite{he2016deep}                         &  $\mathbf{61.2}$                    & \multicolumn{1}{c|}{$\mathbf{69.2}$}     & \multicolumn{1}{c|}{$\mathbf{56.7}$}       & $\mathbf{62.3}$  \\
\bottomrule
\bottomrule
\end{tabular}
\end{table}
\vspace{-1em}
\noindent
A deeper backbone with higher capacity performs better on seen categories, which directly translates to improved performance on the unseen objects. This is largely due to the deeper model learning superior projection matrices as the backbone is able to provide richer feature representations.

\noindent
\textbf{Takeaway. } Backbone capacity directly correlates to improved performance.

\vspace{-1.2em}
\subsubsection{Semantic Embedding Source.}
\vspace{-0.5em}
Semantic embeddings $\mathbf{E}$ for object categories can be derived from different sources such as lingual data (GloVe \cite{pennington2014glove}, Word2Vec \cite{mikolov2013efficient}, ConceptNet \cite{speer2017conceptnet}, and SBERT \cite{reimers2019sentence}), and visuo-lingual information (CLIP \cite{radford2021learning}). Owing to the difference in embedding quality and characteristics, the performance of zero-shot methods is directly impacted by this choice. We train our model with different semantic embeddings to further examine the effects of this decision.
\vspace{-1em}
\begin{table}[H]
\setlength{\aboverulesep}{0pt}
\setlength{\belowrulesep}{0pt}
\setlength{\extrarowheight}{.75ex}
\setlength{\tabcolsep}{4.2pt}
\centering
\begin{tabular}{c|c|c|ccc}
\toprule
\toprule
\rowcolor{Gray}
 &  &  & \multicolumn{3}{c}{GZSD}\\ \cline{4-6} 
\rowcolor{Gray}
\multirow{-2}{*}{Metric} & \multirow{-2}{*}{Embedding} & \multirow{-2}{*}{ZSD} & \multicolumn{1}{c|}{Seen} & \multicolumn{1}{c|}{Unseen} & HM \\ \hline
\multirow{5}{*}{mAP}    & GloVe \cite{pennington2014glove}                   &  $13.8$                    & \multicolumn{1}{c|}{$47.0$}     & \multicolumn{1}{c|}{$8.9$}       &  $15.0$   \\ 
                        & Word2Vec \cite{mikolov2013efficient}                           &  $13.9$                    & \multicolumn{1}{c|}{$47.3$}     & \multicolumn{1}{c|}{$9.4$}       & $15.7$  \\ 
                        & SBERT \cite{reimers2019sentence}                           &     $13.7$                 & \multicolumn{1}{c|}{$46.8$}     & \multicolumn{1}{c|}{$9.8$}       & $16.2$   \\
                        & ConceptNet \cite{speer2017conceptnet}                           &           $15.0$           & \multicolumn{1}{c|}{$\mathbf{47.5}$}     & \multicolumn{1}{c|}{$10.4$}       &  $17.1$  \\ 
                        & CLIP \cite{radford2021learning}                           &                   $\mathbf{17.0}$   & \multicolumn{1}{c|}{$47.0$}     & \multicolumn{1}{c|}{$\mathbf{12.7}$}       & $\mathbf{20.0}$   \\ \cdashline{1-6}
\multirow{5}{*}{Recall}    & GloVe \cite{pennington2014glove}                          &    $59.9$                  & \multicolumn{1}{c|}{$68.2$}     & \multicolumn{1}{c|}{$55.4$}       & $61.1$   \\
                        & Word2Vec \cite{mikolov2013efficient}                           &      $59.7$                & \multicolumn{1}{c|}{$\mathbf{68.5}$}     & \multicolumn{1}{c|}{$55.1$}       &  $61.1$  \\ 
                        & SBERT \cite{reimers2019sentence}                           &   $59.1$                   & \multicolumn{1}{c|}{$67.5$}     & \multicolumn{1}{c|}{$54.9$}       &  $60.6$  \\
                        & ConceptNet \cite{speer2017conceptnet}                           &     $\mathbf{59.9}$                 & \multicolumn{1}{c|}{$68.3$}     & \multicolumn{1}{c|}{$\mathbf{55.7}$}       &  $\mathbf{61.4}$  \\ 
                        & CLIP \cite{radford2021learning}                           &                   $59.2$   & \multicolumn{1}{c|}{$66.9$}     & \multicolumn{1}{c|}{$55.1$}       &  $60.4$  \\ 
\bottomrule
\bottomrule
\end{tabular}
\end{table}
\vspace{-1em}
\noindent
The experiment highlights the considerable impact of embedding choice on model performance, wherein encodings obtained from richer sources like ConceptNet \cite{speer2017conceptnet} that leverages knowledge graphs, or CLIP \cite{radford2021learning} that is trained on both visual and lingual data, provide better performance.

\noindent
\textbf{Takeaway. }Richer category embeddings are better able to facilitate transfer -- identify and localize unseen categories.
\vspace{-0.5em}
\subsubsection{Formulation of Background Embeddings}
\vspace{-0.5em}
As the task of ZSD require models to accurately localize unseen objects, being able to distinguish them from no-category (background) objects is of paramount importance. Existing works either assign a static embedding to the background categories \cite{bansal2018zero, rahman2018zero}, or learn a background embedding using seen category annotations \cite{zheng2020background,zheng2021zero}. To analyze this further, we experiment with three kinds of background embeddings -- \circled{i}
a fixed embedding $[1, . . . , 0]$ as in \cite{bansal2018zero}, 
\circled{ii} 
average over the seen category embeddings $\frac{1}{|\mathcal{C}^s|} \sum_{\mathbf{e}_s \in \mathbf{E}^s} \mathbf{e}_s$ as in \cite{rahman2018zero}, and \circled{iii}
an embedding $\mathbf{b}$ learned alongside the projection matrices described in Section \ref{sec:approach}.
\vspace{-0.5em}
\begin{table}[H]
\setlength{\aboverulesep}{0pt}
\setlength{\belowrulesep}{0pt}
\setlength{\extrarowheight}{.75ex}
\setlength{\tabcolsep}{4.5pt}
\centering
\begin{tabular}{c|c|c|ccc}
\toprule
\toprule
\rowcolor{Gray}
 &  &  & \multicolumn{3}{c}{GZSD}\\ \cline{4-6} 
\rowcolor{Gray}
\multirow{-2}{*}{Metric} & \multirow{-2}{*}{\begin{tabular}[c]{@{}c@{}}Background\\ Embedding\end{tabular}} & \multirow{-2}{*}{ZSD} & \multicolumn{1}{c|}{Seen} & \multicolumn{1}{c|}{Unseen} & HM \\ \hline
\multirow{3}{*}{mAP}    & Fixed \cite{bansal2018zero}                          &                $13.6$      & \multicolumn{1}{c|}{$47.2$}     & \multicolumn{1}{c|}{$9.3$}       &  $15.5$  \\
                        & Mean \cite{rahman2018zero}                           &  $13.6$                    & \multicolumn{1}{c|}{$\mathbf{47.4}$}     & \multicolumn{1}{c|}{$9.1$}       &   $15.3$ \\
                        & Learned (Ours)                            &  $\mathbf{13.9}$                    & \multicolumn{1}{c|}{$47.3$}     & \multicolumn{1}{c|}{$\mathbf{9.4}$}       &   $\mathbf{15.7}$ \\\cdashline{1-6}
\multirow{3}{*}{Recall}    & Fixed \cite{bansal2018zero}                          &  $59.3$                & \multicolumn{1}{c|}{$67.7$}     & \multicolumn{1}{c|}{$54.9$}       &  $60.6$  \\
                        & Mean \cite{rahman2018zero}                           &  $55.7$                        & \multicolumn{1}{c|}{$\mathbf{68.6}$}     & \multicolumn{1}{c|}{$50.6$}       &   $58.2$ \\
                        & Learned (Ours)                               &      $\mathbf{59.7}$                & \multicolumn{1}{c|}{$68.5$}     & \multicolumn{1}{c|}{$\mathbf{55.1}$}       &  $\mathbf{61.1}$  \\
\bottomrule
\bottomrule
\end{tabular}
\end{table}
\vspace{-1em}
\noindent
Compared to learning a background embedding on the seen class information, having a static background embedding (fixed vector or mean) provides inferior performance on the unseen categories. This can primarily be attributed to the static embeddings not being independent of the training data, and therefore not being able to accurately distinguish background from the unseen category objects. 

\noindent
\textbf{Takeaway. }Learning a background embedding is preferable to using a static background embedding.
\vspace{-1em}
\subsubsection{Formulation of Regressor}
\label{sec:regressor}
\vspace{-0.6em}
Effective localization of unseen objects is heavily conditioned on the quality of unseen category regressors. Existing works have looked at heuristically utilizing the seen category regressors as a proxy for their unseen counterparts \cite{hayat2020synthesizing, khandelwal2021unit}, or leveraging a semantic space projection from image features to the embedding space \cite{rahman2018polarity,zheng2021zero}. Here we explore the impact of the type of transfer used, by comparing four variants of our proposed model with different formulations for regressor transfer -- \circled{i} Using no transfer, and directly using the bounding box predicted by the RPN without any refinement, \circled{ii} Using the most similar seen category regressor as a proxy for its unseen counterpart, \circled{iii} Using a linear combination of seen category regressor outputs based on embedding similarity between $\mathbf{E}^s$ and $\mathbf{E}^u$, and \circled{iv} Using our proposed transfer described in Section \ref{sec:approach}.
\vspace{-1em}
\begin{table}[H]
\setlength{\aboverulesep}{0pt}
\setlength{\belowrulesep}{0pt}
\setlength{\extrarowheight}{.75ex}
\setlength{\tabcolsep}{3pt}
\centering
\begin{tabular}{c|c|c|ccc}
\toprule
\toprule
\rowcolor{Gray}
 &  &  & \multicolumn{3}{c}{GZSD}\\ \cline{4-6} 
\rowcolor{Gray}
\multirow{-2}{*}{Metric} & \multirow{-2}{*}{\begin{tabular}[c]{@{}c@{}}Regressor\\ Formulation\end{tabular}} & \multirow{-2}{*}{ZSD} & \multicolumn{1}{c|}{Seen} & \multicolumn{1}{c|}{Unseen} & HM \\ \hline
\multirow{4}{*}{mAP}    & No Transfer                         &   $8.3$                    & \multicolumn{1}{c|}{$47.4$}     & \multicolumn{1}{c|}{$5.9$}       &  $10.5$  \\
                        & Most Similar                            &      $13.9$                & \multicolumn{1}{c|}{$47.3$}     & \multicolumn{1}{c|}{$9.4$}       &  $15.7$  \\
                        & Linear Combination                            &  $13.7$                     & \multicolumn{1}{c|}{$\mathbf{47.4}$}     & \multicolumn{1}{c|}{$9.3$}       & $15.5$\\
                        & Learned (Ours)                           &                        $\mathbf{13.9}$                    & \multicolumn{1}{c|}{$47.3$}     & \multicolumn{1}{c|}{$\mathbf{9.4}$}       &   $\mathbf{15.7}$ \\ \cdashline{1-6}
\multirow{4}{*}{Recall}    & No Transfer                         &   $56.3$                   & \multicolumn{1}{c|}{$68.6$}     & \multicolumn{1}{c|}{$50.9$}       &  $58.4$  \\
                        & Most Similar                            & $59.4$                      & \multicolumn{1}{c|}{$68.5$}     & \multicolumn{1}{c|}{$54.6$}       & $60.8$   \\
                        & Linear Combination                            &    $59.7$                  & \multicolumn{1}{c|}{$\mathbf{68.6}$}     & \multicolumn{1}{c|}{$54.9$}       & $61.0$ \\
                        & Learned (Ours)                           &    $\mathbf{59.7}$                  & \multicolumn{1}{c|}{$68.5$}     & \multicolumn{1}{c|}{$\mathbf{55.1}$}       &  $\mathbf{61.1}$  \\
\bottomrule
\bottomrule
\end{tabular}
\end{table}
\vspace{-1.2em}

\noindent
Forgoing regressor information transfer leads to a drastic decline in performance. Heuristic transfer mechanisms like selecting the most similar or taking a linear combination over the seen category regressors, although provide a significant improvement over the no transfer approach, are slightly inferior to our proposed learned transfer. For the unseen categories, the values generated by these heuristics are constrained to be smaller than or equal to the most similar seen category activation. The learned transfer circumvents this restriction, leading to better performance. Learning the transfer for the segmentor has similar benefits. The segmentor formulation ablations are shown in the \textbf{appendix}.

\noindent
\textbf{Takeaway. }A learned transfer mechanism outperforms heuristic approaches.

\vspace{-1em}
\subsubsection{Fine-Tuning Parameters}
\vspace{-0.5em}
When learning the projection matrices, our proposed approach freezes the learnable parameters of the Faster-RCNN \cite{ren2015faster} (or Mask-RCNN \cite{he2017mask}) architecture. To justify this decision, we train two variants of our approach wherein these learnable parameters are also updated alongside the projection matrices, namely -- \circled{i} training only the proposal-level feature extractor, and \circled{ii} training all the learnable parameters. The results are shown in the table below.
\vspace{-0.5em}
\begin{table}[H]
\setlength{\aboverulesep}{0pt}
\setlength{\belowrulesep}{0pt}
\setlength{\extrarowheight}{.75ex}
\setlength{\tabcolsep}{1.7pt}
\centering
\begin{tabular}{c|c|c|ccc}
\toprule
\toprule
\rowcolor{Gray}
 &  &  & \multicolumn{3}{c}{GZSD}\\ \cline{4-6} 
\rowcolor{Gray}
\multirow{-2}{*}{Metric} & \multirow{-2}{*}{\begin{tabular}[c]{@{}c@{}}Unfrozen\\ Parameters\end{tabular}} & \multirow{-2}{*}{ZSD} & \multicolumn{1}{c|}{Seen} & \multicolumn{1}{c|}{Unseen} & HM \\ \hline
\multirow{3}{*}{mAP}    &   All                       &    $12.4$                  & \multicolumn{1}{c|}{$47.0$}     & \multicolumn{1}{c|}{$8.7$}       & $14.7$   \\
                        & Proposal Feat. Extractor                             &    $12.8$                  & \multicolumn{1}{c|}{$\mathbf{47.6}$}     & \multicolumn{1}{c|}{$8.8$}       &  $14.9$  \\
                        & None (Ours)                           &                        $\mathbf{13.9}$                    & \multicolumn{1}{c|}{$47.3$}     & \multicolumn{1}{c|}{$\mathbf{9.4}$}       &  $\mathbf{15.7}$  \\\cdashline{1-6}
\multirow{3}{*}{Recall}    &   All                       & $58.2$                     & \multicolumn{1}{c|}{$67.7$}     & \multicolumn{1}{c|}{$54.0$}       & $60.1$   \\
                        & Proposal Feat. Extractor                             &        $59.4$              & \multicolumn{1}{c|}{$68.2$}     & \multicolumn{1}{c|}{$54.8$}       &   $60.8$ \\
                        & None (Ours)           &    $\mathbf{59.7}$                  & \multicolumn{1}{c|}{$\mathbf{68.5}$}     & \multicolumn{1}{c|}{$\mathbf{55.1}$}       & $\mathbf{61.1}$    \\
\bottomrule
\bottomrule
\end{tabular}
\end{table}
\vspace{-1em}
\noindent
It can be seen that training the learnable parameters alongside the projection matrices is sub-optimal, and leads to the model overfitting on the seen categories. 

\noindent
\textbf{Takeaway. }Freezing the learnable parameters of Faster/Mask RCNN when learning the projection matrices enables better generalization to unseen category objects.

\vspace{-1em}
\subsubsection{Classifier Formulation and Loss}
\vspace{-0.5em}
With regards to the proposal-level classifier, our proposed approach, described in Section \ref{sec:approach}, uses a simple matrix product between \emph{normalized} category embeddings and the projected proposal-level features. The aforementioned classifier is trained with a cross-entropy loss. Existing methods have explored other classifier formulations, that in turn dictate the type of loss employed during training. Bansal \etal~\cite{bansal2018zero} use cosine similarity based classifier with a max-margin loss, whereas \cite{gupta2020multi} compute an L2-error between the projected image features and category embeddings within their semantic model. We ablate these choices and contrast the performance obtained in the table below.
\vspace{-1em}
\begin{table}[H]
\setlength{\aboverulesep}{0pt}
\setlength{\belowrulesep}{0pt}
\setlength{\extrarowheight}{.75ex}
\setlength{\tabcolsep}{2.7pt}
\centering
\begin{tabular}{c|c|c|ccc}
\toprule
\toprule
\rowcolor{Gray}
 &  &  & \multicolumn{3}{c}{GZSD}\\ \cline{4-6} 
\rowcolor{Gray}
\multirow{-2}{*}{Metric} & \multirow{-2}{*}{\begin{tabular}[c]{@{}c@{}}Classifier\\ Loss\end{tabular}} & \multirow{-2}{*}{ZSD} & \multicolumn{1}{c|}{Seen} & \multicolumn{1}{c|}{Unseen} & HM \\ \hline
\multirow{3}{*}{mAP}    & Max Margin \cite{bansal2018zero}                         &    $12.7$                  & \multicolumn{1}{c|}{$46.3$}     & \multicolumn{1}{c|}{$6.7$}       &  $11.7$  \\
                        & L2 Error \cite{gupta2020multi}                            &  $12.0$                    & \multicolumn{1}{c|}{$39.8$}     & \multicolumn{1}{c|}{$5.6$}       & $9.8$   \\
                        & Cross Entropy (Ours)                           &                        $\mathbf{13.9}$                    & \multicolumn{1}{c|}{$\mathbf{47.3}$}     & \multicolumn{1}{c|}{$\mathbf{9.4}$}       &   $\mathbf{15.7}$ \\ \cdashline{1-6}
\multirow{3}{*}{Recall}    & Max Margin \cite{bansal2018zero}                         & $54.3$                     & \multicolumn{1}{c|}{$67.3$}     & \multicolumn{1}{c|}{$49.1$}       &   $60.1$ \\
                        & L2 Error \cite{gupta2020multi}                            &   $48.6$                   & \multicolumn{1}{c|}{$62.1$}     & \multicolumn{1}{c|}{$34.2$}       & $44.1$   \\
                        & Cross Entropy (Ours)                           &    $\mathbf{59.7}$                  & \multicolumn{1}{c|}{$\mathbf{68.5}$}     & \multicolumn{1}{c|}{$\mathbf{55.1}$}       &  $\mathbf{61.1}$   \\
\bottomrule
\bottomrule
\end{tabular}
\end{table}
\vspace{-1em}
\noindent
It is evident that using a max-margin or L2-error based loss to train the classifier provides inferior identification of unseen category objects. The cross-entropy loss is consistent with the formulation used in Faster-RCNN \cite{ren2015faster} (or Mask-RCNN \cite{he2017mask}) to train the classifier, and therefore provides better performance. Additionally, unlike the max-margin loss that relies on the selection of a appropriate margin, the cross-entropy loss has no such hyperparameter.

\noindent
\textbf{Takeaway. }Cross-entropy based formulation is easier to train and provides better performance.

\vspace{-1em}
\subsubsection{Seen-Unseen Performance Trade-off}
\vspace{-0.5em}
We use a threshold $\beta$ to bias the model towards unseen categories while simultaneously forgoing the need for re-training. We further explore the impact of this biasing by evaluating our trained model on the GZSD setup with different $\beta$ values, and visualize the results in the figure below (mAP on the left, recall on the right). Note that the ZSD setup is not affected by the choice of $\beta$ as the seen category predictions are simply ignored (\ie $\beta > 1$).
\begin{figure}[H]
\centering
\begin{subfigure}[b]{0.5\linewidth}
   \includegraphics[width=\linewidth]{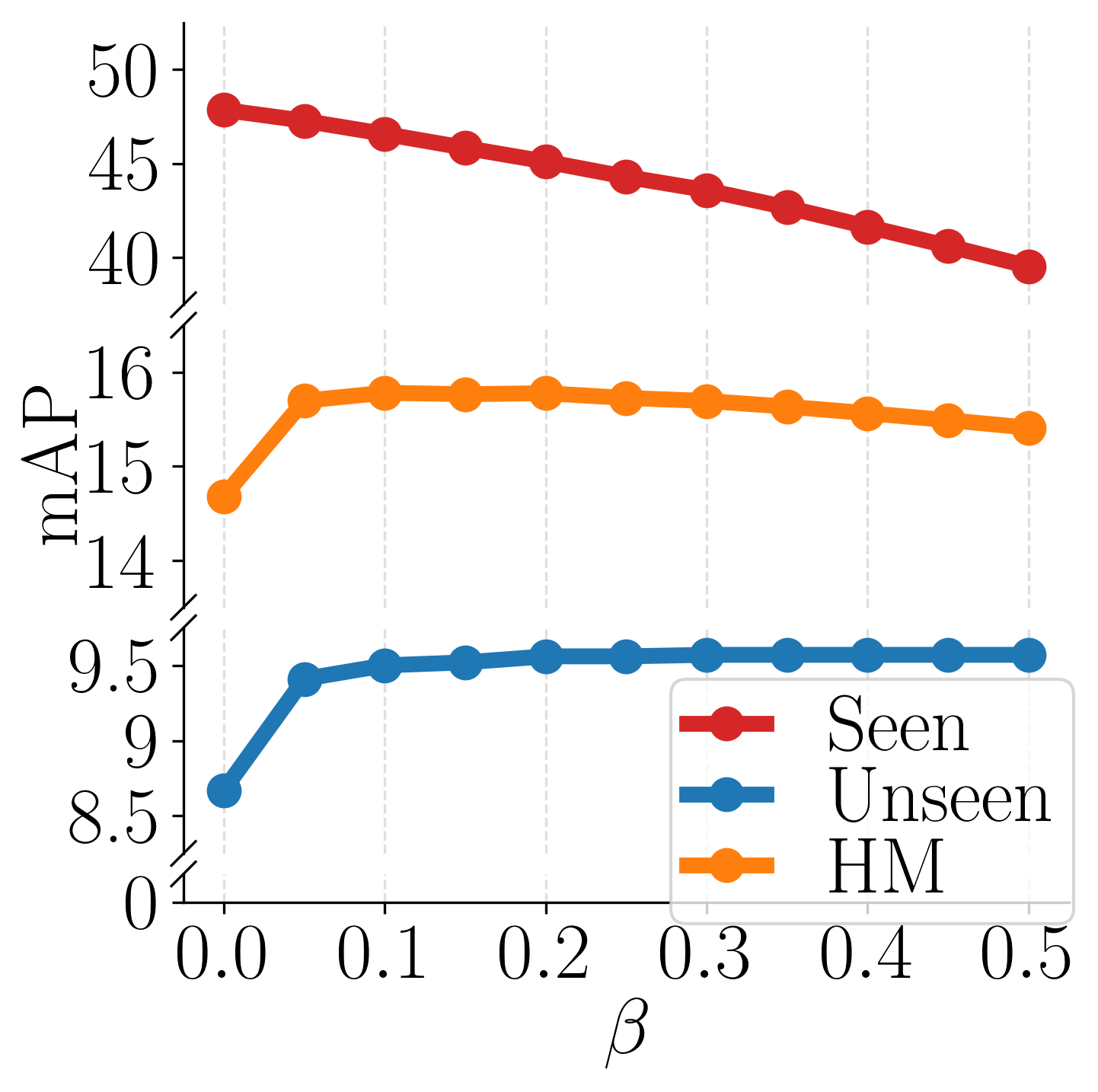}
   \label{fig:Ng1} 
\end{subfigure}\begin{subfigure}[b]{0.5\linewidth}
   \includegraphics[width=\linewidth]{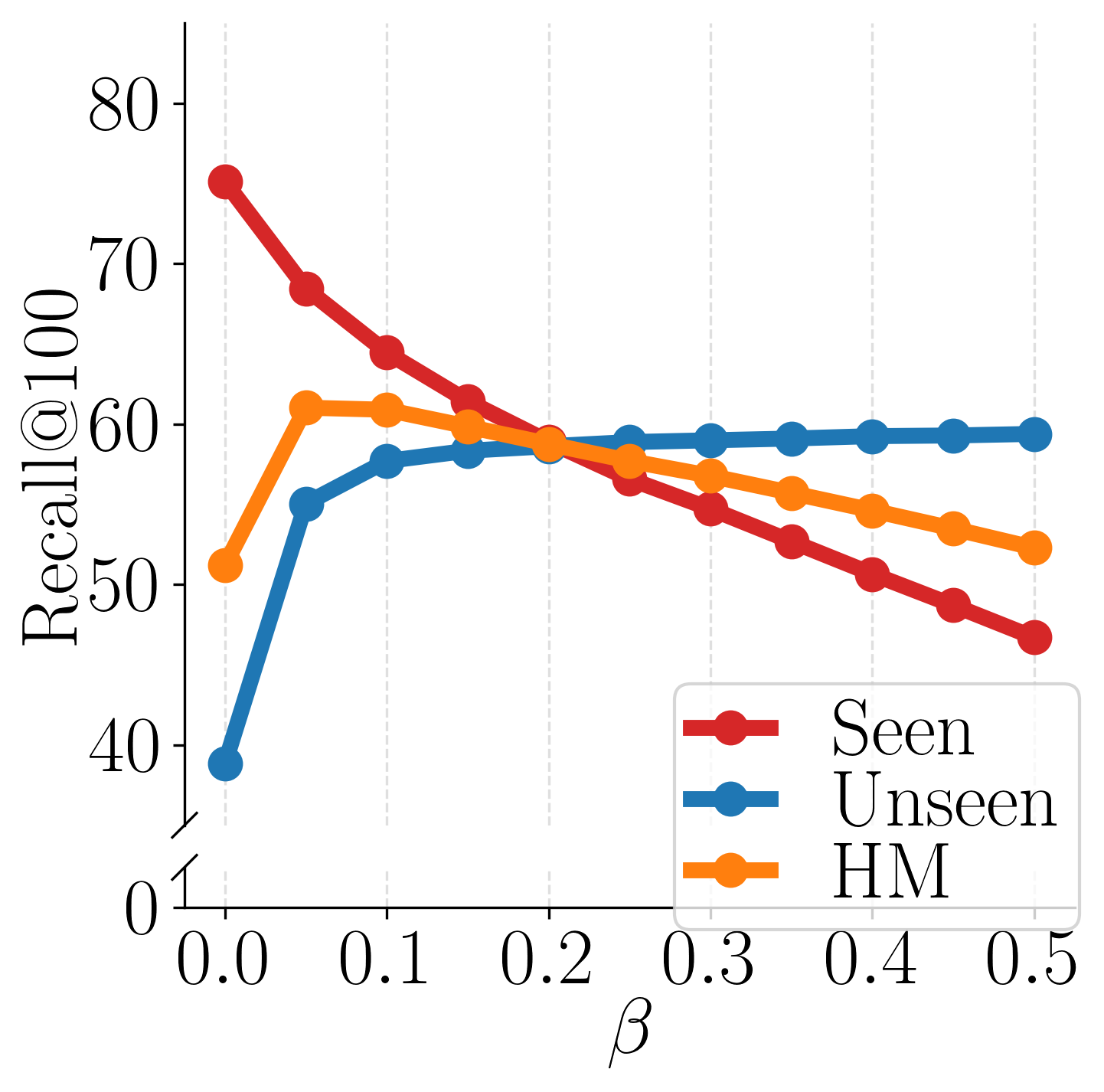}
   \label{fig:Ng2}
\end{subfigure}
\vspace{-4em}
\end{figure}
\noindent
Higher values of $\beta$ generally leads to an increase in unseen category performance at the expense of its seen category counterpart. These gains, however, plateau beyond $\beta=0.2$, and further increasing $\beta$ only hurts the seen category performance. Therefore, we find $\beta=0.05$ to provide a reasonable trade-off between seen-unseen category performance.

\noindent
\textbf{Takeaway. }An appropriate $\beta$ greatly boosts unseen category performance without the need for re-training.

\begin{table}[t]
\setlength{\aboverulesep}{0pt}
\setlength{\belowrulesep}{0pt}
\setlength{\extrarowheight}{.75ex}
\setlength{\tabcolsep}{5.5pt}
\centering
\begin{tabular}{cc|c|ccc|c}
\toprule
\toprule
\rowcolor{Gray}
& &  & \multicolumn{3}{c|}{Recall$@100$}                                  & mAP    \\ \cline{3-6} 
\rowcolor{Gray}
                  &      &                        & \multicolumn{3}{c|}{IoU}                                           & IoU    \\ \cline{3-6} 
\rowcolor{Gray}
 \multicolumn{2}{c|}{\multirow{-3}{*}{Method}}                       &  \multirow{-3}{*}{Split}                      & \multicolumn{1}{c|}{$0.4$}  & \multicolumn{1}{c|}{$0.5$}  & $0.6$  & $0.5$  \\ \hline
\multicolumn{1}{c|}{\multirow{8}{*}{\rotatebox[origin=c]{90}{F-RCNN}}} & SB \cite{bansal2018zero}                      & $48/17$                & \multicolumn{1}{c|}{$34.5$}             & \multicolumn{1}{c|}{$22.1$}             & $11.3$             & $0.3$              \\
\multicolumn{1}{c|}{}& DSES \cite{bansal2018zero}                    & $48/17$                & \multicolumn{1}{c|}{$40.2$}             & \multicolumn{1}{c|}{$27.2$}             & $13.6$             & $0.5$              \\
\multicolumn{1}{c|}{}& TD \cite{li2019zero}                     & $48/17$                & \multicolumn{1}{c|}{$45.5$}             & \multicolumn{1}{c|}{$34.3$}             & $18.1$             & $-$                \\
\multicolumn{1}{c|}{}& PL \cite{rahman2018polarity}                     & $48/17$                & \multicolumn{1}{c|}{$-$}                & \multicolumn{1}{c|}{$43.5$}             & $-$                & $10.1$             \\
\multicolumn{1}{c|}{}& BLC  \cite{zheng2020background}                   & $48/17$                & \multicolumn{1}{c|}{$51.3$}             & \multicolumn{1}{c|}{$48.8$}             & $45.0$             & $10.6$             \\
\multicolumn{1}{c|}{}& RRFS  \cite{huang2022robust}                  & $48/17$                & \multicolumn{1}{c|}{$58.1$}             & \multicolumn{1}{c|}{$53.5$}             & $47.9$             & $13.4$             \\
\multicolumn{1}{c|}{}& Ours$_{\text{F} {+} \text{W2V}}$                     &  $48/17$                       & \multicolumn{1}{c|}{\textcolor{red}{$65.7$}}             & \multicolumn{1}{c|}{\textcolor{red}{$61.2$}}             &    \textcolor{blue}{$54.5$}         & \textcolor{blue}{$14.8$}
\\
\multicolumn{1}{c|}{}& Ours$_{\text{F} {+} \text{CLIP}}$                     &  $48/17$                       & \multicolumn{1}{c|}{\textcolor{blue}{$64.9$}}             & \multicolumn{1}{c|}{\textcolor{blue}{$61.0$}}             &    \textcolor{red}{$54.6$}         & \textcolor{red}{$18.4$}             \\
\cdashline{1-7}
\multicolumn{1}{c|}{\multirow{3}{*}{\rotatebox[origin=c]{90}{M-RCNN}}}& ZSI  \cite{zheng2021zero}                   & $48/17$               & \multicolumn{1}{c|}{$57.4$}              & \multicolumn{1}{c|}{$53.9$}             &   $48.3$          & $11.4$             \\
\multicolumn{1}{c|}{}& Ours$_{\text{M} {+} \text{W2V}}$                     &  $48/17$                       & \multicolumn{1}{c|}{\textcolor{red}{$66.5$}}             & \multicolumn{1}{c|}{\textcolor{red}{$62.7$}}             & \textcolor{red}{$56.2$}           & \textcolor{blue}{$15.3$} \\   
\multicolumn{1}{c|}{}& Ours$_{\text{M} {+} \text{CLIP}}$                     &  $48/17$                       & \multicolumn{1}{c|}{\textcolor{blue}{$66.0$}}             & \multicolumn{1}{c|}{\textcolor{blue}{$62.5$}}             &    \textcolor{blue}{$55.7$}         & \textcolor{red}{$18.2$}             \\
\hline
\multicolumn{1}{c|}{\multirow{6}{*}{\rotatebox[origin=c]{90}{F-RCNN}}}& PL \cite{rahman2018polarity}                     & $65/15$                & \multicolumn{1}{c|}{$-$}                & \multicolumn{1}{c|}{$37.7$}             & $-$                & $12.4$             \\
\multicolumn{1}{c|}{}& BLC \cite{zheng2020background}                     & $65/15$                & \multicolumn{1}{c|}{$57.2$}             & \multicolumn{1}{c|}{$54.7$}             & $51.2$             & $14.7$             \\
\multicolumn{1}{c|}{}& SU \cite{hayat2020synthesizing}                      & $65/15$                & \multicolumn{1}{c|}{$54.4$}             & \multicolumn{1}{c|}{$54.0$}             & $47.0$             & $19.0$             \\
\multicolumn{1}{c|}{}& RRFS \cite{huang2022robust}                    & $65/15$                & \multicolumn{1}{c|}{$65.3$}             & \multicolumn{1}{c|}{$62.3$}             & $55.9$             & \textcolor{blue}{$19.8$}   \\
\multicolumn{1}{c|}{}& Ours$_{\text{F} {+} \text{W2V}}$                     &  $65/15$                       & \multicolumn{1}{c|}{\textcolor{blue}{$74.7$}}             & \multicolumn{1}{c|}{\textcolor{blue}{$72.1$}}             &    \textcolor{blue}{$66.6$}         & $19.6$    
\\
\multicolumn{1}{c|}{}& Ours$_{\text{F} {+} \text{CLIP}}$                     &  $65/15$                       & \multicolumn{1}{c|}{\textcolor{red}{$76.9$}}             & \multicolumn{1}{c|}{\textcolor{red}{$73.8$}}             &    \textcolor{red}{$68.7$}         & \textcolor{red}{$24.5$}             \\
\cdashline{1-7}
\multicolumn{1}{c|}{\multirow{3}{*}{\rotatebox[origin=c]{90}{M-RCNN}}}& ZSI \cite{zheng2021zero}                     & $65/15$               & \multicolumn{1}{c|}{$61.9$}              & \multicolumn{1}{c|}{$58.9$}             &   $54.4$          & $13.6$             \\
\multicolumn{1}{c|}{}& Ours$_{\text{M} {+} \text{W2V}}$                       &  $65/15$                       & \multicolumn{1}{c|}{\textcolor{blue}{$75.2$}}             &  \multicolumn{1}{c|}{\textcolor{blue}{$71.9$}}             & \textcolor{blue}{$66.4$}            & \textcolor{blue}{$19.0$}             \\
\multicolumn{1}{c|}{}& Ours$_{\text{M} {+} \text{CLIP}}$                     &  $65/15$                       & \multicolumn{1}{c|}{\textcolor{red}{$76.7$}}             & \multicolumn{1}{c|}{\textcolor{red}{$73.8$}}            &    \textcolor{red}{$68.5$}         &    \textcolor{red}{$25.1$}         \\
\bottomrule
\bottomrule
\end{tabular}
\vspace{-0.8em}
\caption{\textbf{Zero-Shot Detection (ZSD).} mAP at $\text{IoU}{=}0.5$, Recall$@100$ at $\text{IoU}{=}[0.4,0.5,0.6]$ reported for unseen categories. Best result highlighted in \textcolor{red}{red}, second best in \textcolor{blue}{blue}.   }
\label{tab:zsd}
\vspace{-2em}
\end{table}

\begin{table}[t]
\setlength{\aboverulesep}{0pt}
\setlength{\belowrulesep}{0pt}
\setlength{\extrarowheight}{.75ex}
\setlength{\tabcolsep}{0.9pt}
\centering
\begin{tabular}{cc|c|cc|cc|cc}
\toprule
\toprule
\rowcolor{Gray}
 & &  \multirow{2}{*}{} & \multicolumn{2}{c|}{Seen} & \multicolumn{2}{c|}{Unseen} & \multicolumn{2}{c}{HM} \\ \cline{4-9} 
\rowcolor{Gray}
                 \multicolumn{2}{c|}{\multirow{-2}{*}{Method}}              &   \multirow{-2}{*}{Split}                     & mAP         & Recall      & mAP          & Recall       & mAP        & Recall    \\ \hline
\multicolumn{1}{c|}{\multirow{5}{*}{\rotatebox[origin=c]{90}{F-RCNN}}} &   PL \cite{rahman2018polarity}                    & $48/17$                & $35.9$      & $38.2$      & $4.1$        & $26.3$       & $7.4$      & $31.2$    \\
\multicolumn{1}{c|}{}  &   BLC \cite{zheng2020background}                     & $48/17$                & $42.1$      & $57.6$      & $4.5$        & $46.4$       & $8.2$      & $51.4$    \\
\multicolumn{1}{c|}{}  &   RRFS \cite{huang2022robust}                    & $48/17$                & $42.3$      & $59.7$      & \textcolor{blue}{$13.4$}       & \textcolor{red}{$58.8$}       & \textcolor{blue}{$20.4$}     & $59.2$    \\
\multicolumn{1}{c|}{}  &   Ours$_{\text{F} {+} \text{W2V}}$                      & $48/17$                & \textcolor{red}{$48.9$}            & \textcolor{red}{$69.2$}            & $10.2$             & \textcolor{blue}{$56.7$}              & $16.9$           & \textcolor{red}{$62.3$}          \\
\multicolumn{1}{c|}{}  &   Ours$_{\text{F} {+} \text{CLIP}}$                      & $48/17$                & \textcolor{blue}{$48.6$}            & \textcolor{blue}{$69.2$}            & \textcolor{red}{$13.9$}             & $56.4$              &    \textcolor{red}{$21.6$}        &  \textcolor{blue}{$62.1$}         \\ \cdashline{1-9}
\multicolumn{1}{c|}{\multirow{3}{*}{\rotatebox[origin=c]{90}{M-RCNN}}} &   ZSI \cite{zheng2021zero}                     & $48/17$                & $46.5$      & \textcolor{red}{$70.8$}      & $4.8$        & $53.9$       & $8.8$      & $61.2$    \\
\multicolumn{1}{c|}{}  &   Ours$_{\text{M} {+} \text{W2V}}$                      & $48/17$                & \textcolor{red}{$49.5$}            &    \textcolor{blue}{$70.7$}         &    \textcolor{blue}{$10.6$}          & \textcolor{blue}{$58.0$}              & \textcolor{blue}{$17.5$}           &   \textcolor{red}{$63.7$}        \\
\multicolumn{1}{c|}{}  &   Ours$_{\text{M} {+} \text{CLIP}}$                      & $48/17$                & \textcolor{blue}{$49.4$}            & $69.8$            & \textcolor{red}{$13.6$}             &          \textcolor{red}{$58.3$}     &    \textcolor{red}{$21.3$}        & \textcolor{blue}{$63.5$}          \\
\hline
\multicolumn{1}{c|}{\multirow{6}{*}{\rotatebox[origin=c]{90}{F-RCNN}}}&   PL \cite{rahman2018polarity}                    & $65/15$                & $34.1$      & $36.4$      & $12.4$       & $37.2$       & $18.2$     & $36.8$    \\
\multicolumn{1}{c|}{}&   BLC \cite{zheng2020background}                    & $65/15$                & $36.0$      & $56.4$      & $13.1$       & $51.7$       & $19.2$     & $53.9$    \\
\multicolumn{1}{c|}{}&   SU \cite{hayat2020synthesizing}                     & $65/15$                & $36.9$      & $57.7$      & $19.0$       & $53.9$       & $25.1$     & $55.8$    \\
\multicolumn{1}{c|}{}&   RRFS \cite{huang2022robust}                   & $65/15$                & $37.4$      & $58.6$      & \textcolor{blue}{$19.8$}       & $61.8$       & $26.0$     & $60.2$    \\
\multicolumn{1}{c|}{}&   Ours$_{\text{F} {+} \text{W2V}}$                      & $65/15$                & \textcolor{blue}{$40.2$}            & \textcolor{blue}{$70.8$}            & $19.3$             & \textcolor{blue}{$64.2$}              &  \textcolor{blue}{$26.1$}          &  \textcolor{blue}{$67.3$}         \\
\multicolumn{1}{c|}{}&   Ours$_{\text{F} {+} \text{CLIP}}$                      & $65/15$                & \textcolor{red}{$40.3$}            & \textcolor{red}{$70.9$}            & \textcolor{red}{$24.2$}             & \textcolor{red}{$66.6$}              &   \textcolor{red}{$30.2$}         & \textcolor{red}{$68.7$}          \\ \cdashline{1-9}
\multicolumn{1}{c|}{\multirow{3}{*}{\rotatebox[origin=c]{90}{M-RCNN}}}&   ZSI \cite{zheng2021zero}                    & $65/15$                & $38.7$      & $67.1$      & $13.6$       & $58.9$       & $20.1$     & $62.8$    \\
\multicolumn{1}{c|}{}&   Ours$_{\text{M} {+} \text{W2V}}$                       & $65/15$                &   \textcolor{blue}{$40.7$}          & \textcolor{blue}{$70.0$}            & \textcolor{blue}{$18.8$}             & \textcolor{blue}{$64.6$}              &   \textcolor{blue}{$25.7$}         & \textcolor{blue}{$67.2$}     \\
\multicolumn{1}{c|}{}&   Ours$_{\text{M} {+} \text{CLIP}}$                      & $65/15$                & \textcolor{red}{$40.9$}            & \textcolor{red}{$70.0$}            & \textcolor{red}{$24.9$}             & \textcolor{red}{$66.5$}              &    \textcolor{red}{$30.9$}        &  \textcolor{red}{$68.2$}  \\
\bottomrule
\bottomrule
\end{tabular}
\vspace{-0.8em}
\caption{\textbf{Generalized Zero-Shot Detection (GZSD).} mAP, Recall$@100$, and the harmonic mean (HM) between seen and unseen category performance is reported at $\text{IoU}{=}0.5$. Best result highlighted in \textcolor{red}{red}, second best in \textcolor{blue}{blue}. }
\label{tab:gzsd}
\vspace{-2.5em}
\end{table}

\vspace{-0.5em}
\section{Experiments}
\vspace{-0.5em}
\label{sec:experiment}
We compare our proposed model, described in Section \ref{sec:approach}, which has carefully constructed using the best performing design components (Section \ref{sec:designchoices}), against existing methods. 

\vspace{0.2em}
\noindent
\textbf{Dataset. }The evaluation is done on the MSCOCO 2014 \cite{lin2014microsoft} dataset, which contains $82,783$ training images and $40,504$ validation images with $80$ categories. 

\vspace{0.2em}
\noindent
\textbf{Seen-Unseen Splits. }For the task of ZSD, consistent with existing work in \cite{bansal2018zero, rahman2018polarity}, we report performance on two seen-unseen category splits -- \circled{i} the $48/17$ split \cite{bansal2018zero}, and \circled{ii} the $65/15$ split \cite{rahman2018polarity}. For the task of ZSI, we adopt the $48/17$ and $65/15$ splits proposed in \cite{zheng2021zero}, Following the setup in \cite{bansal2018zero,rahman2018polarity,zheng2021zero}, for each task and split, we remove \emph{all} images containing unseen categories from the training set to guarantee that unseen objects will not influence model training.

\vspace{0.2em}
\noindent
\textbf{Evaluation.} Following existing work, we report performance on the standard MSCOCO metrics, namely mean average precision (mAP) at $\text{IoU}{=}0.5$ and recall@100 at three different IoU thresholds $[0.4, 0.5, 0.6]$. For the GZSD/GZSI tasks, we also compute the harmonic mean (HM) between the seen and unseen category performance.

\vspace{0.2em}
\noindent
\textbf{Implementation Details.} To enable fair comparison with recent methods \cite{hayat2020synthesizing,huang2022robust,zheng2021zero}, we train our proposed approach on the ResNet-101 \cite{he2016deep} backbone. The learnable parameters of Faster RCNN \cite{ren2015faster} / Mask RCNN \cite{he2017mask} are trained on the seen category annotations using the SGD optimizer for $100,000$ iterations with a batch size of $16$ and a learning rate of $0.02$, which is decayed by a factor of $0.1$ at $60,000$ and $80,000$ iterations. These parameters are frozen during the second training step, and the projection matrices are estimated using a learning rate of $0.005$. During inference, for each image, our model generates $100$ predictions. \textcolor{blue}{We will make our code and all models public upon acceptance.}

\vspace{-0.5em}
\subsection{Comparison to Existing Methods}
\label{sec:sota-comparison}
\vspace{-0.5em}
We report performance using: \circled{i} a lingual embedding in Word2Vec \cite{mikolov2013efficient}, denoted as ``W2V", and \circled{ii} a visio-lingual embedding in CLIP \cite{mikolov2013efficient}. The Faster-RCNN \cite{ren2015faster} architecture, denoted as ``F'', is used for ZSD/GZSD tasks. Similarly, the Mask-RCNN \cite{he2017mask} architecture, denoted as ``M'', is used for the ZSI/GZSI tasks. We differentiate variants of our method by the architecture and embedding choice. For example, ``$\text{F}+\text{W2V}$'' represents the use of Faster-RCNN \cite{ren2015faster} architecture with Word2Vec \cite{mikolov2013efficient} embeddings. 

\vspace{0.2em}
\noindent
\textbf{Zero-Shot Detection. }Comparisons to existing methods on the ZSD setup are shown in Table \ref{tab:zsd}. For the $48/17$ split, our ``$\text{F}{+}\text{W2V}$'' variant provides $10.4\%$ higher mAP, and on average $13.7\%$ higher recall across the three thresholds when compared with the most competitive method in \cite{huang2022robust}. The difference on mAP is more pronounced with the use of a richer embedding source, wherein our ``$\text{F}{+}\text{CLIP}$'' variant achieves a $37.3\%$ improvement on mAP with $13.1\%$ increase on recall over the method in \cite{huang2022robust}. A similar observation holds for the $65/15$ split, where the ``$\text{F}{+}\text{CLIP}$'' variant outperforms the closest baseline \cite{huang2022robust} by $23.7\%$ and $19.6\%$ on mAP and recall respectively. Under the more challenging GZSD setup, as highlighted in Table \ref{tab:gzsd}, our variants on average provide $5\%$ and $12.9\%$ higher HM recall on the $47/17$ and $65/15$ splits respectively. Although our ``$\text{F}{+}\text{W2V}$'' variant has a slightly worse performance when compared to RRFS \cite{huang2022robust} on HM mAP, the ``$\text{F}{+}\text{CLIP}$'' variant has an average improvement of $11\%$ on HM mAP, demonstrating the ability of our simplistic approach to effectively detect both seen and unseen objects simultaneously. 

\vspace{0.2em}
\noindent
\textbf{Zero-Shot Segmentation.} Comparisons to the baseline in \cite{zheng2021zero} on the ZSI and GZSI tasks are presented in Tables \ref{tab:zsi} and \ref{tab:gzsi} respectively. For the ZSI task, irrespective of the embedding choice, we outperform the closest baseline in \cite{zheng2021zero} by $84.9\%$ and $31.3\%$ on mAP and recall respectively, on average, across the two splits. Similar improvements are seen on the GZSI task, wherein our model variants provide an average increase of $117.4\%$ and $12.8\%$ on HM mAP and HM Recall respectively over \cite{zheng2021zero} across the two splits, highlighting the superior performance of our approach on both seen and unseen category segmentation.

\vspace{0.2em}
\noindent
\textbf{Additional Results.} Qualitative visualisations and per-category results are shown in the \textbf{appendix}.

\begin{table}[t]
\setlength{\aboverulesep}{0pt}
\setlength{\belowrulesep}{0pt}
\setlength{\extrarowheight}{.75ex}
\setlength{\tabcolsep}{7pt}
\centering
\begin{tabular}{c|c|ccc|c}
\toprule
\toprule
\rowcolor{Gray}
 &  & \multicolumn{3}{c|}{Recall$@100$}                                  & mAP    \\ \cline{3-6} 
\rowcolor{Gray}
                        &                        & \multicolumn{3}{c|}{IoU}                                           & IoU    \\ \cline{3-6} 
\rowcolor{Gray}
 \multirow{-3}{*}{Method}                       &  \multirow{-3}{*}{Split}                      & \multicolumn{1}{c|}{$0.4$}  & \multicolumn{1}{c|}{$0.5$}  & $0.6$  & $0.5$  \\ \hline
 
ZSI  \cite{zheng2021zero}                   & $48/17$               & \multicolumn{1}{c|}{$50.3$}             & \multicolumn{1}{c|}{$44.9$}             &   $38.7$          & $9.0$             \\
Ours$_{\text{M} {+} \text{W2V}}$                     &  $48/17$                       & \multicolumn{1}{c|}{\textcolor{blue}{$62.1$}}             & \multicolumn{1}{c|}{\textcolor{blue}{$56.7$}}             & \textcolor{blue}{$49.1$}           & \textcolor{blue}{$14.0$} \\  
Ours$_{\text{M} {+} \text{CLIP}}$                     &  $48/17$                       & \multicolumn{1}{c|}{\textcolor{red}{$62.1$}}             &  \multicolumn{1}{c|}{\textcolor{red}{$56.9$}}             & \textcolor{red}{$49.8$}          & \textcolor{red}{$17.1$}   \\    
\hline
ZSI \cite{zheng2021zero}                     & $65/15$               & \multicolumn{1}{c|}{$55.8$}              & \multicolumn{1}{c|}{$50.0$}             &   $42.9$          & $10.5$             \\
Ours$_{\text{M} {+} \text{W2V}}$                     &  $65/15$                       & \multicolumn{1}{c|}{\textcolor{blue}{$71.8$}}             & \multicolumn{1}{c|}{\textcolor{blue}{$67.5$}}             & \textcolor{blue}{$61.6$}            & \textcolor{blue}{$17.9$}            \\
Ours$_{\text{M} {+} \text{CLIP}}$                     &  $65/15$                       & \multicolumn{1}{c|}{\textcolor{red}{$73.2$}}             &  \multicolumn{1}{c|}{\textcolor{red}{$68.6$}}             & \textcolor{red}{$62.4$}          & \textcolor{red}{$23.5$}   \\    
\bottomrule
\bottomrule
\end{tabular}
\vspace{-0.8em}
\caption{\textbf{Zero-Shot Segmentation (ZSI).} mAP at $\text{IoU}{=}0.5$, Recall$@100$ at $\text{IoU}{=}[0.4,0.5,0.6]$ reported for unseen categories. Best result highlighted in \textcolor{red}{red}, second best in \textcolor{blue}{blue}.}
\label{tab:zsi}
\vspace{-1em}
\end{table}

\begin{table}[t]
\setlength{\aboverulesep}{0pt}
\setlength{\belowrulesep}{0pt}
\setlength{\extrarowheight}{.75ex}
\setlength{\tabcolsep}{1.5pt}
\centering
\begin{tabular}{c|c|cc|cc|cc}
\toprule
\toprule
\rowcolor{Gray}
 &  & \multicolumn{2}{c|}{Seen} & \multicolumn{2}{c|}{Unseen} & \multicolumn{2}{c}{HM} \\ \cline{3-8} 
\rowcolor{Gray}
          \multirow{-2}{*}{Method}              &   \multirow{-2}{*}{Split}                     & mAP         & Recall      & mAP          & Recall       & mAP        & Recall    \\ \hline
ZSI \cite{zheng2021zero}                    & $48/17$                & $43.0$      & $64.5$      & $3.7$        & $44.9$       & $6.7$      & $52.9$    \\
Ours$_{\text{M} {+} \text{W2V}}$                      & $48/17$                &  \textcolor{red}{$46.3$}           &    \textcolor{red}{$65.7$}         &    \textcolor{blue}{$9.7$}          &    \textcolor{blue}{$52.5$}           &  \textcolor{blue}{$16.0$}          & \textcolor{blue}{$58.4$}          \\
Ours$_{\text{M} {+} \text{CLIP}}$                      & $48/17$                & \textcolor{blue}{$46.2$}            & \textcolor{blue}{$65.2$}            & \textcolor{red}{$12.9$}             & \textcolor{red}{$53.1$}               &   \textcolor{red}{$20.2$}         &    \textcolor{red}{$58.5$}       \\
\hline
ZSI \cite{zheng2021zero}                     & $65/15$                & $35.8$      & $62.6$      & $10.6$       & $50.0$       & $16.2$     & $55.6$    \\
Ours$_{\text{M} {+} \text{W2V}}$                       & $65/15$                &   \textcolor{blue}{$38.7$}          & \textcolor{blue}{$65.9$}            & \textcolor{blue}{$17.7$}             & \textcolor{blue}{$61.3$}              & \textcolor{blue}{$24.3$}           & \textcolor{blue}{$63.5$}     \\
Ours$_{\text{M} {+} \text{CLIP}}$                      & $65/15$                & \textcolor{red}{$38.7$}            & \textcolor{red}{$65.9$}            & \textcolor{red}{$23.3$}             & \textcolor{red}{$63.5$}               &  \textcolor{red}{$29.1$}          &    \textcolor{red}{$64.7$}       \\
\bottomrule
\bottomrule
\end{tabular}
\vspace{-0.8em}
\caption{\textbf{Generalized Zero-Shot Segmentation (GZSI).} mAP, Recall$@100$, and the harmonic mean (HM) between seen and unseen category performance is reported at $\text{IoU}{=}0.5$. Best result highlighted in \textcolor{red}{red}, second best in \textcolor{blue}{blue}.}
\label{tab:gzsi}
\vspace{-2em}
\end{table}

\vspace{-0.5em}
\section{Conclusion}
\vspace{-0.5em}
In this work we present a simple approach to zero-shot detection and segmentation that is carefully constructed through extensive ablations over critical design choices. Through extensive experimentation we highlight the superior performance of our method when compared to more complex architectures, and suggest the need to revisit some of the recent design trends in the ZSD/ZSI field, wherein our method can act as a strong baseline.

{\small
\bibliographystyle{ieee_fullname}
\bibliography{egbib}
}

\clearpage

\appendix

\clearpage
\twocolumn[
\begin{center}
    {\LARGE \textbf{Appendix}}
    \vspace{2em}
\end{center}
\vspace{1em}
]

\section{Formulation of Segmentor}
In Section \ref{sec:regressor} of the main paper, we demonstrate the advantages of our proposed learned regressor formulation when compared to heuristics defined over seen categories. In this section we highlight that learning the transfer for the unseen category segmentors has similar benefits by comparing it against their heuristic counterparts. Specifically, we use a Mask RCNN \cite{he2017mask} based ResNet-50 \cite{he2016deep} variant of our model proposed in Section \ref{sec:approach} trained using the Word2Vec embeddings \cite{mikolov2013efficient}, and compare between four different formulations for segmentor transfer -- \circled{i} Using no transfer, and predicting a zero mask, \circled{ii} Using the most similar seen category segmentor as a proxy for its unseen counterpart, \circled{iii} Using a linear combination of seen category segmentor outputs based on embedding similarity between $\mathbf{E}^s$ and $\mathbf{E}^u$, and \circled{iv} Using our proposed segmentor transfer described in Section \ref{sec:approach}. The performance of these variants is reported on the ZSI/GZSI tasks using the $48/17$ split in the table below.

\begin{table}[H]
\setlength{\aboverulesep}{0pt}
\setlength{\belowrulesep}{0pt}
\setlength{\extrarowheight}{.75ex}
\setlength{\tabcolsep}{3pt}
\centering
\begin{tabular}{c|c|c|ccc}
\toprule
\toprule
\rowcolor{Gray}
 &  &  & \multicolumn{3}{c}{GZSI}\\ \cline{4-6} 
\rowcolor{Gray}
\multirow{-2}{*}{Metric} & \multirow{-2}{*}{\begin{tabular}[c]{@{}c@{}}Segmentor\\ Formulation\end{tabular}} & \multirow{-2}{*}{GZSI} & \multicolumn{1}{c|}{Seen} & \multicolumn{1}{c|}{Unseen} & HM \\ \hline
\multirow{4}{*}{mAP}    & No Transfer                         &   $0.0$                    & \multicolumn{1}{c|}{$45.1$}     & \multicolumn{1}{c|}{$0.0$}       & $0.0$  \\
                        & Most Similar                            &      $13.0$                & \multicolumn{1}{c|}{$45.0$}     & \multicolumn{1}{c|}{$8.8$}       & $14.7$  \\
                        & Linear Combination                            &  $12.6$                     & \multicolumn{1}{c|}{$\mathbf{45.1}$}     & \multicolumn{1}{c|}{$8.5$}       & $14.3$ \\
                        & Learned (Ours)                           &                        $\mathbf{13.3}$                    & \multicolumn{1}{c|}{$44.9$}     & \multicolumn{1}{c|}{$\mathbf{8.9}$}       &  $\mathbf{14.9}$  \\ \cdashline{1-6}
\multirow{4}{*}{Recall}    & No Transfer                         &   $0.0$                   & \multicolumn{1}{c|}{$65.1$}     & \multicolumn{1}{c|}{$0.0$}       & $0.0$   \\
                        & Most Similar                            & $54.3$                      & \multicolumn{1}{c|}{$65.1$}     & \multicolumn{1}{c|}{$49.8$}       & $56.4$  \\
                        & Linear Combination                            &    $53.5$                  & \multicolumn{1}{c|}{$\mathbf{65.1}$}     & \multicolumn{1}{c|}{$49.0$}       & $55.9$ \\
                        & Learned (Ours)                           &    $\mathbf{55.7}$                  & \multicolumn{1}{c|}{$64.9$}     & \multicolumn{1}{c|}{$\mathbf{51.1}$}       &  $\mathbf{57.2}$  \\
\bottomrule
\bottomrule
\end{tabular}
\end{table}
It can be seen that not using any segmentor information transfer and predicting a zero-mask is unable to produce segmentation masks for the unseen categories. Heuristic transfer mechanisms like selecting the most similar or taking a linear combination over the seen category segmentors, although provide a significant improvement over the no transfer approach, are inferior to our proposed learned transfer. 

\section{Additional Results}
\subsection{Per Category Results}
Please refer to Section \ref{sec:sota-comparison} of the main paper for our model definitions. We report the per-category detection mAP performance for our model variants on the ZSD task for the $48/17$ split in Table \ref{tab:perclass4817} and for the $65/15$ split in Table \ref{tab:perclass6515}. Similarly, the per-category segmentation mAP performance for our model variants on the ZSI task are shown in Tables \ref{tab:perclassseg4817} and \ref{tab:perclassseg6515}. Note that the work in \cite{zheng2021zero} does not provide per-category mAP numbers for the MSCOCO dataset and is therefore excluded from the tables.

\subsection{Qualitative Visualization}
\begin{figure*}[t]
\captionsetup[subfigure]{labelformat=empty}
\begin{subfigure}{.33\textwidth}
\centering
\includegraphics[width=\textwidth, height=4cm]{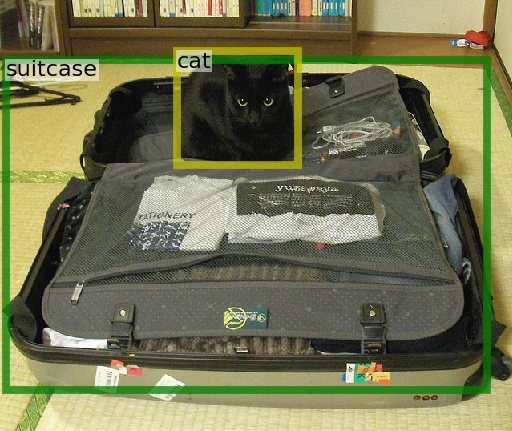}
\end{subfigure}%
\begin{subfigure}{.33\textwidth}
\centering
\includegraphics[width=\textwidth, height=4cm]{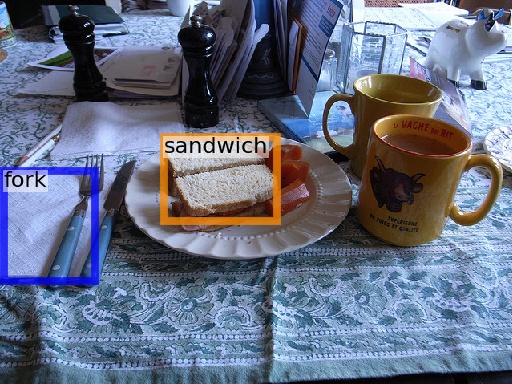}
\end{subfigure}%
\begin{subfigure}{.33\textwidth}
\centering
\includegraphics[width=\textwidth, height=4cm]{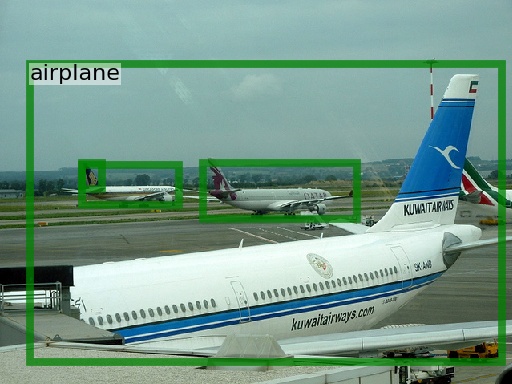}
\end{subfigure}
\begin{subfigure}{.33\textwidth}
\centering
\includegraphics[width=\textwidth, height=4cm]{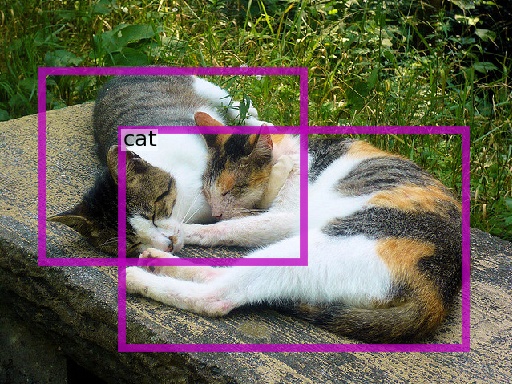}
\end{subfigure}%
\begin{subfigure}{.33\textwidth}
\centering
\includegraphics[width=\textwidth, height=4cm]{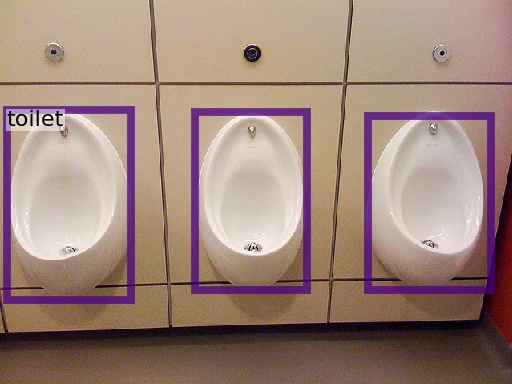}
\end{subfigure}%
\begin{subfigure}{.33\textwidth}
\centering
\includegraphics[width=\textwidth, height=4cm]{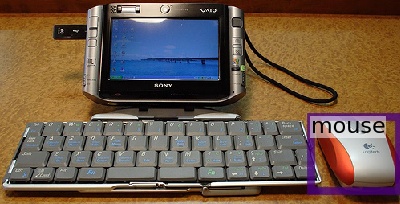}
\end{subfigure}
\begin{subfigure}{.33\textwidth}
\centering
\includegraphics[width=\textwidth, height=4cm]{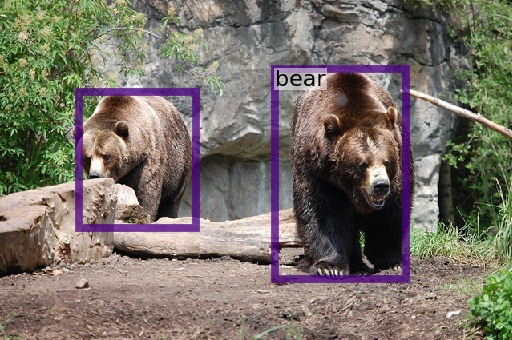}
\end{subfigure}%
\begin{subfigure}{.33\textwidth}
\centering
\includegraphics[width=\textwidth, height=4cm]{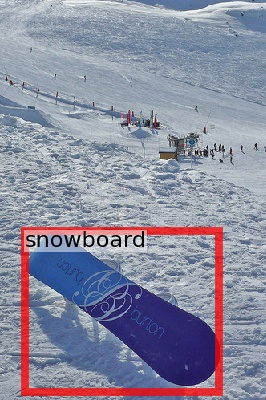}
\end{subfigure}%
\begin{subfigure}{.33\textwidth}
\centering
\includegraphics[width=\textwidth, height=4cm]{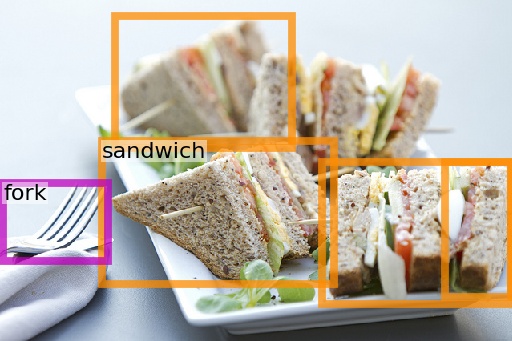}
\end{subfigure}
\begin{subfigure}{.33\textwidth}
\centering
\includegraphics[width=\textwidth, height=4cm]{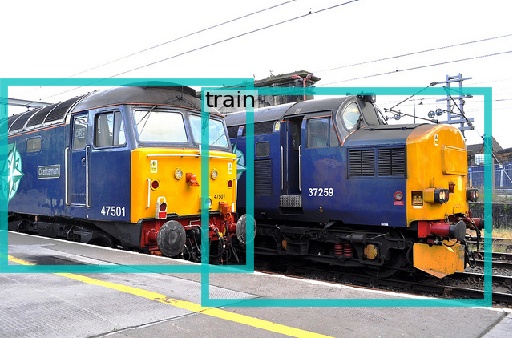}
\end{subfigure}%
\begin{subfigure}{.33\textwidth}
\centering
\includegraphics[width=\textwidth, height=4cm]{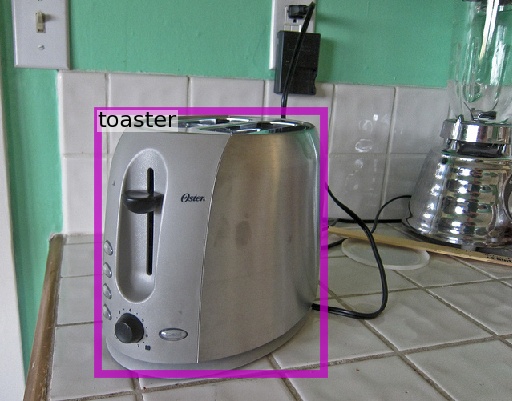}
\end{subfigure}%
\begin{subfigure}{.33\textwidth}
\centering
\includegraphics[width=\textwidth, height=4cm]{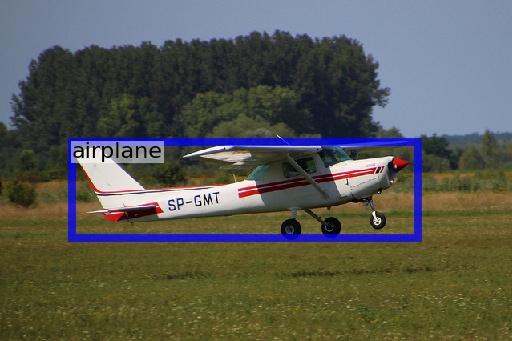}
\end{subfigure}
\caption{\textbf{Qualitative Visualizations.} Unseen category examples for the zero-shot detection task generated using our proposed ``$\text{F}{+}\text{W2V}$'' variant (color $=$ object category).}
\label{fig:qualidetw2v}
\end{figure*}

\begin{figure*}[t]
\captionsetup[subfigure]{labelformat=empty}
\begin{subfigure}{.33\textwidth}
\centering
\includegraphics[width=\textwidth, height=4cm]{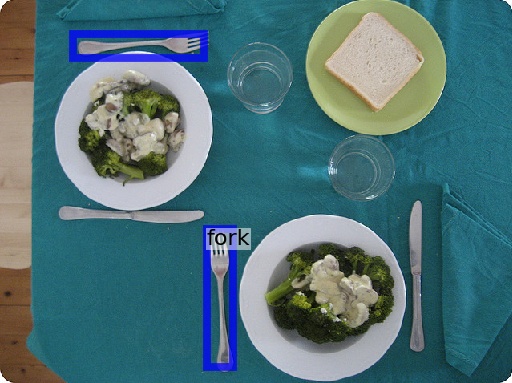}
\end{subfigure}%
\begin{subfigure}{.33\textwidth}
\centering
\includegraphics[width=\textwidth, height=4cm]{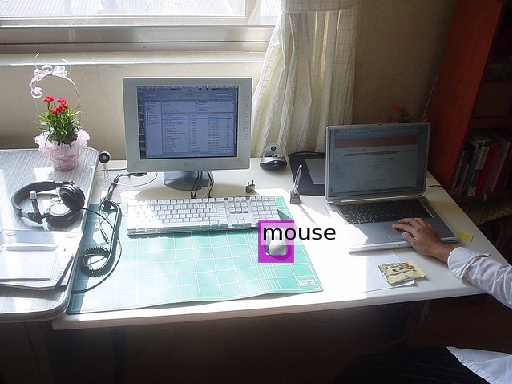}
\end{subfigure}%
\begin{subfigure}{.33\textwidth}
\centering
\includegraphics[width=\textwidth, height=4cm]{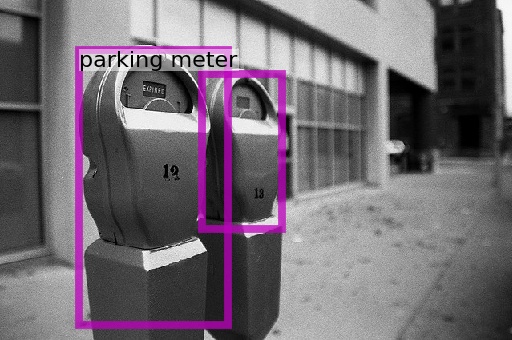}
\end{subfigure}
\begin{subfigure}{.33\textwidth}
\centering
\includegraphics[width=\textwidth, height=4cm]{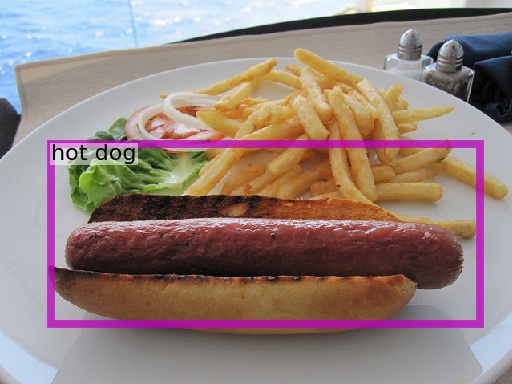}
\end{subfigure}%
\begin{subfigure}{.33\textwidth}
\centering
\includegraphics[width=\textwidth, height=4cm]{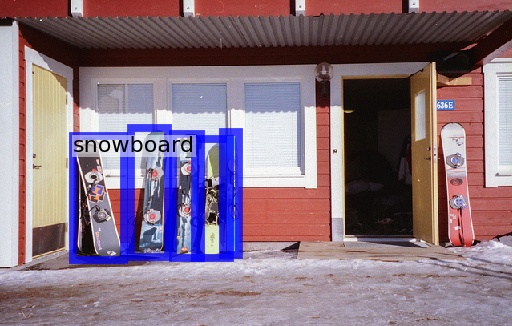}
\end{subfigure}%
\begin{subfigure}{.33\textwidth}
\centering
\includegraphics[width=\textwidth, height=4cm]{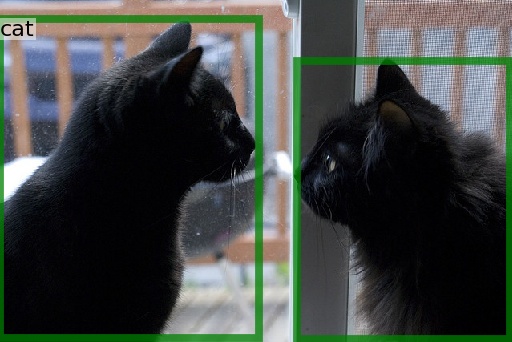}
\end{subfigure}
\begin{subfigure}{.33\textwidth}
\centering
\includegraphics[width=\textwidth, height=4cm]{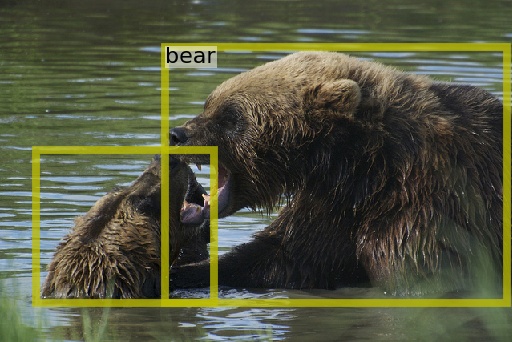}
\end{subfigure}%
\begin{subfigure}{.33\textwidth}
\centering
\includegraphics[width=\textwidth, height=4cm]{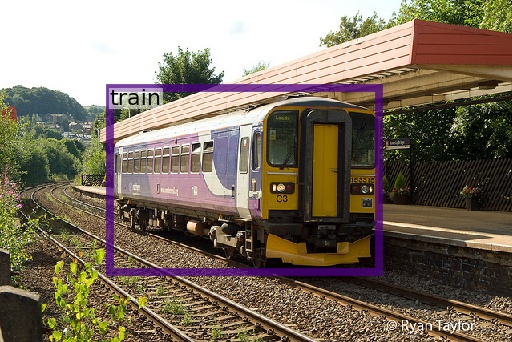}
\end{subfigure}%
\begin{subfigure}{.33\textwidth}
\centering
\includegraphics[width=\textwidth, height=4cm]{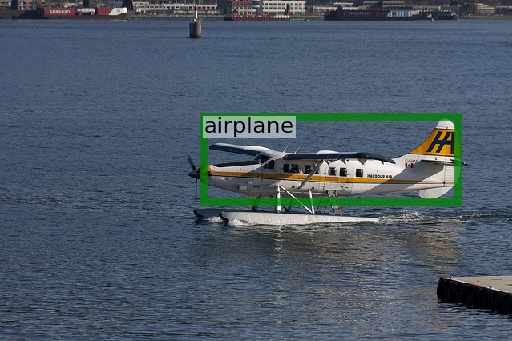}
\end{subfigure}
\begin{subfigure}{.33\textwidth}
\centering
\includegraphics[width=\textwidth, height=4cm]{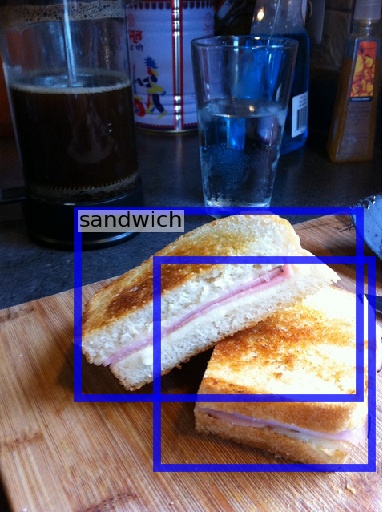}
\end{subfigure}%
\begin{subfigure}{.33\textwidth}
\centering
\includegraphics[width=\textwidth, height=4cm]{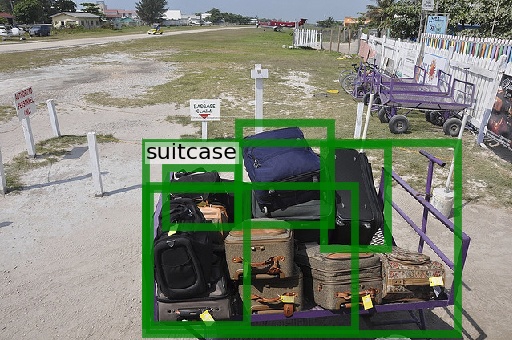}
\end{subfigure}%
\begin{subfigure}{.33\textwidth}
\centering
\includegraphics[width=\textwidth, height=4cm]{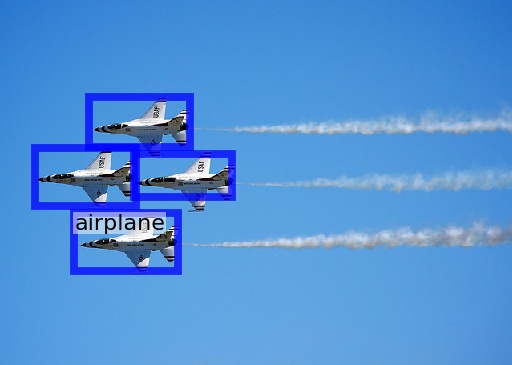}
\end{subfigure}
\caption{\textbf{Qualitative Visualizations.} Unseen category examples for the zero-shot detection task generated using our proposed ``$\text{F}{+}\text{CLIP}$'' variant (color $=$ object category).}
\label{fig:qualidetclip}
\end{figure*}

\begin{figure*}[t]
\captionsetup[subfigure]{labelformat=empty}
\begin{subfigure}{.33\textwidth}
\centering
\includegraphics[width=\textwidth, height=4cm]{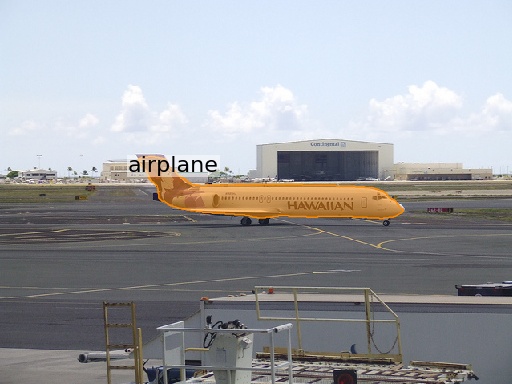}
\end{subfigure}%
\begin{subfigure}{.33\textwidth}
\centering
\includegraphics[width=\textwidth, height=4cm]{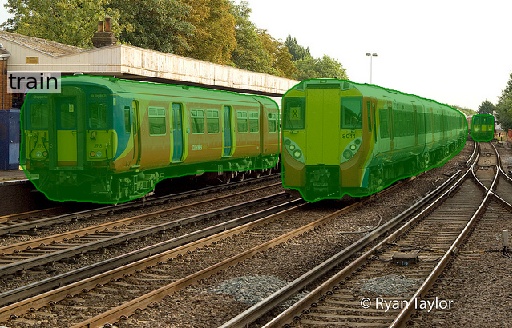}
\end{subfigure}%
\begin{subfigure}{.33\textwidth}
\centering
\includegraphics[width=\textwidth, height=4cm]{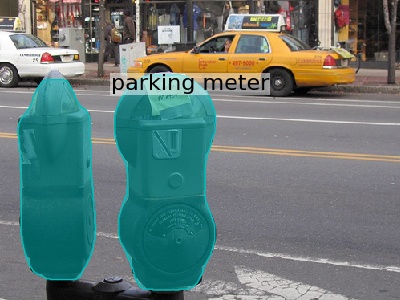}
\end{subfigure}
\begin{subfigure}{.33\textwidth}
\centering
\includegraphics[width=\textwidth, height=4cm]{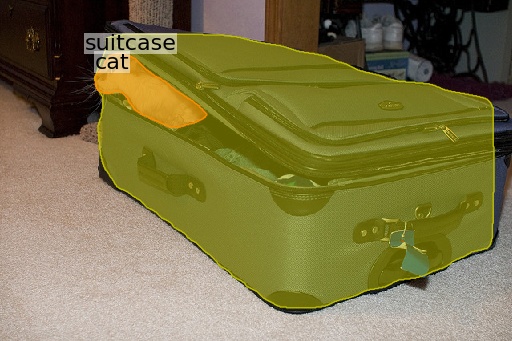}
\end{subfigure}%
\begin{subfigure}{.33\textwidth}
\centering
\includegraphics[width=\textwidth, height=4cm]{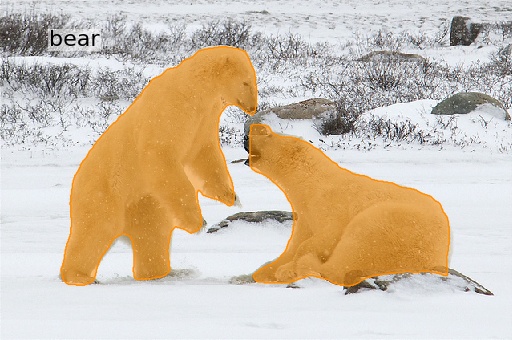}
\end{subfigure}%
\begin{subfigure}{.33\textwidth}
\centering
\includegraphics[width=\textwidth, height=4cm]{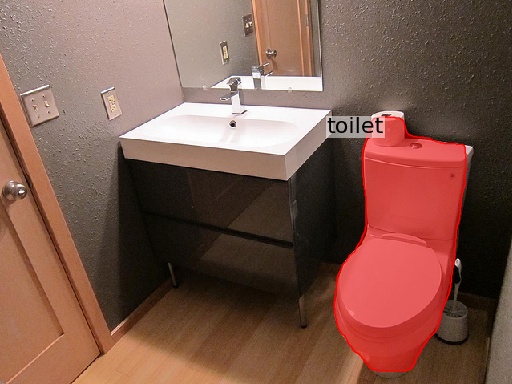}
\end{subfigure}
\begin{subfigure}{.33\textwidth}
\centering
\includegraphics[width=\textwidth, height=4cm]{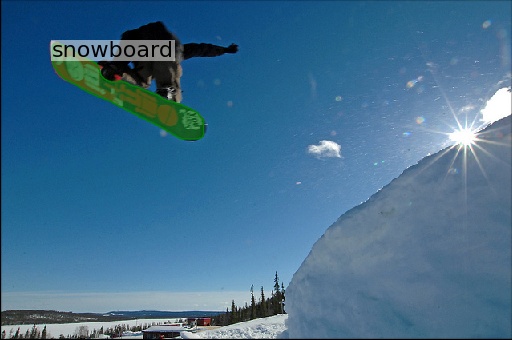}
\end{subfigure}%
\begin{subfigure}{.33\textwidth}
\centering
\includegraphics[width=\textwidth, height=4cm]{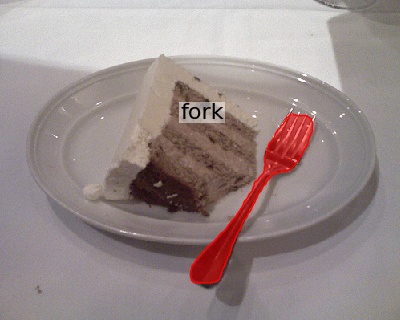}
\end{subfigure}%
\begin{subfigure}{.33\textwidth}
\centering
\includegraphics[width=\textwidth, height=4cm]{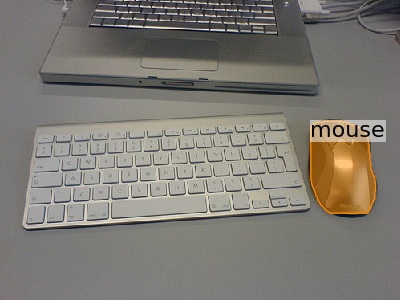}
\end{subfigure}
\begin{subfigure}{.33\textwidth}
\centering
\includegraphics[width=\textwidth, height=4cm]{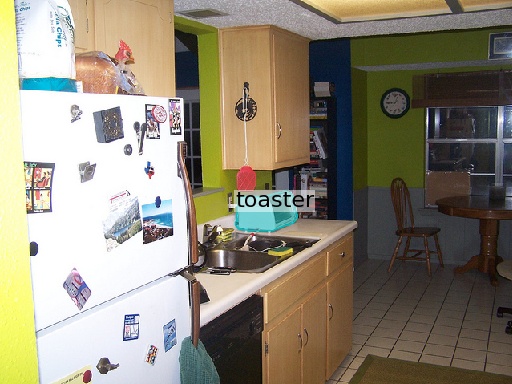}
\end{subfigure}%
\begin{subfigure}{.33\textwidth}
\centering
\includegraphics[width=\textwidth, height=4cm]{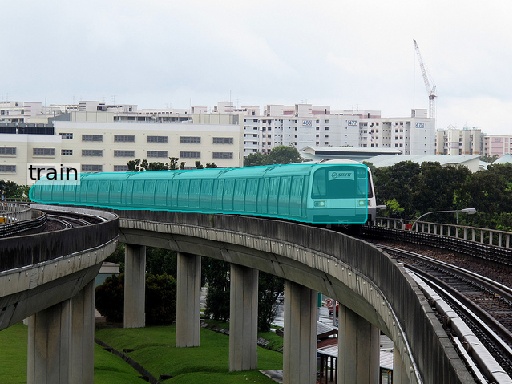}
\end{subfigure}%
\begin{subfigure}{.33\textwidth}
\centering
\includegraphics[width=\textwidth, height=4cm]{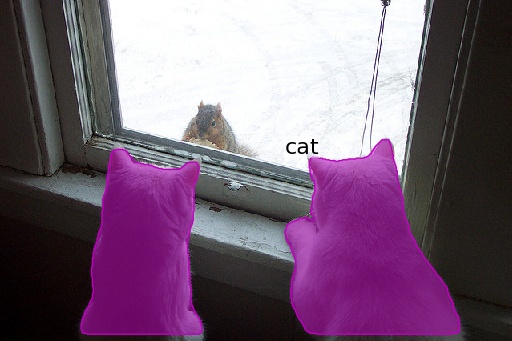}
\end{subfigure}
\caption{\textbf{Qualitative Visualizations.} Unseen category examples for the zero-shot segmentation task generated using our proposed ``$\text{M}{+}\text{W2V}$'' variant (color $=$ object category).}
\label{fig:qualisegw2v}
\end{figure*}

\begin{figure*}[t]
\captionsetup[subfigure]{labelformat=empty}
\begin{subfigure}{.33\textwidth}
\centering
\includegraphics[width=\textwidth, height=4cm]{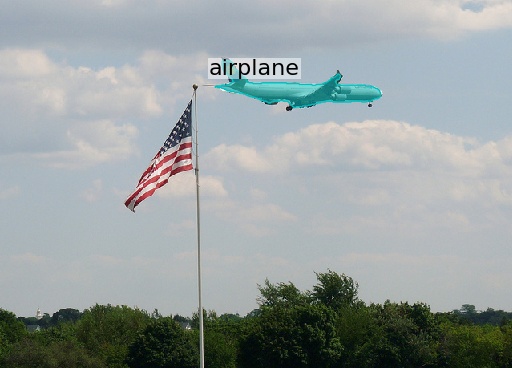}
\end{subfigure}%
\begin{subfigure}{.33\textwidth}
\centering
\includegraphics[width=\textwidth, height=4cm]{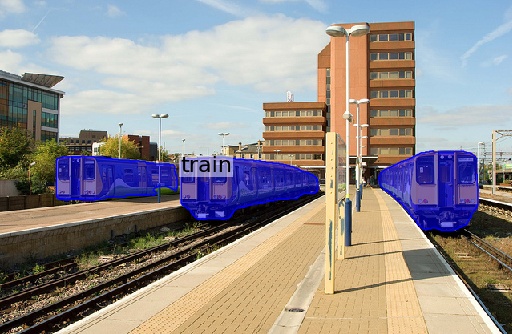}
\end{subfigure}%
\begin{subfigure}{.33\textwidth}
\centering
\includegraphics[width=\textwidth, height=4cm]{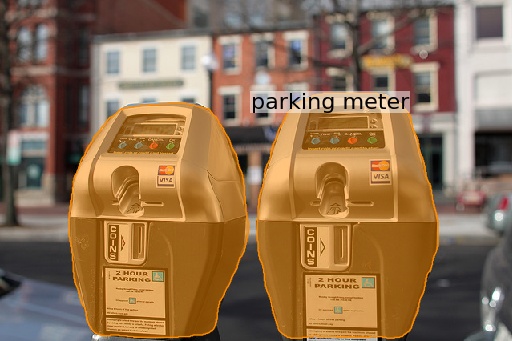}
\end{subfigure}
\begin{subfigure}{.33\textwidth}
\centering
\includegraphics[width=\textwidth, height=4cm]{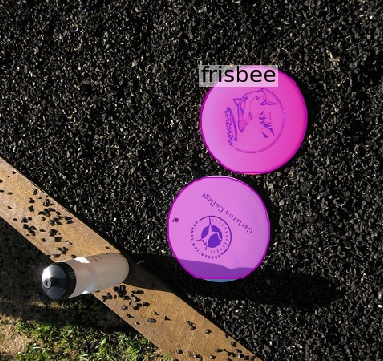}
\end{subfigure}%
\begin{subfigure}{.33\textwidth}
\centering
\includegraphics[width=\textwidth, height=4cm]{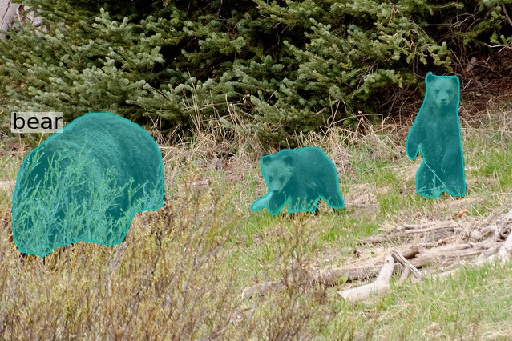}
\end{subfigure}%
\begin{subfigure}{.33\textwidth}
\centering
\includegraphics[width=\textwidth, height=4cm]{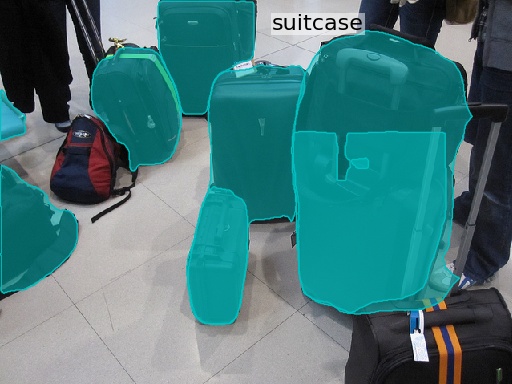}
\end{subfigure}
\begin{subfigure}{.33\textwidth}
\centering
\includegraphics[width=\textwidth, height=4cm]{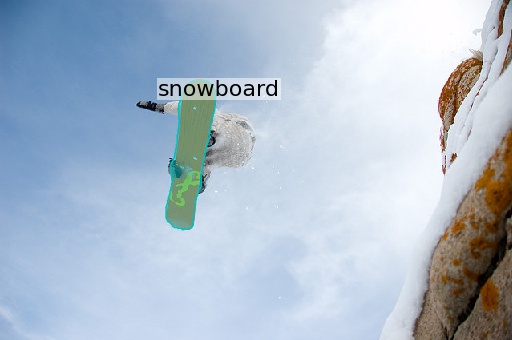}
\end{subfigure}%
\begin{subfigure}{.33\textwidth}
\centering
\includegraphics[width=\textwidth, height=4cm]{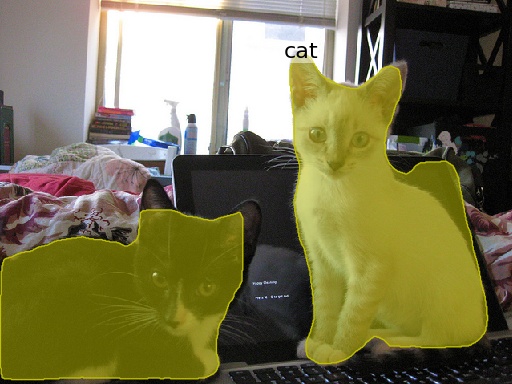}
\end{subfigure}%
\begin{subfigure}{.33\textwidth}
\centering
\includegraphics[width=\textwidth, height=4cm]{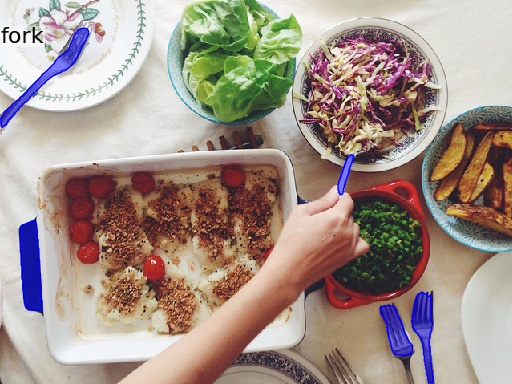}
\end{subfigure}
\begin{subfigure}{.33\textwidth}
\centering
\includegraphics[width=\textwidth, height=4cm]{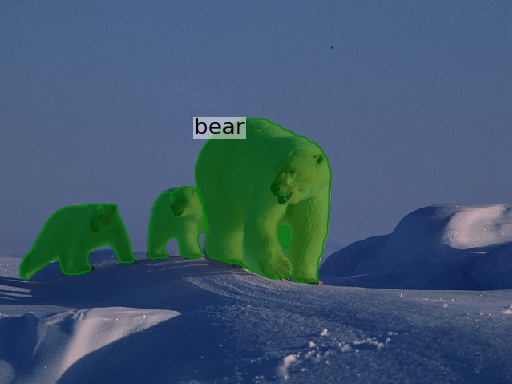}
\end{subfigure}%
\begin{subfigure}{.33\textwidth}
\centering
\includegraphics[width=\textwidth, height=4cm]{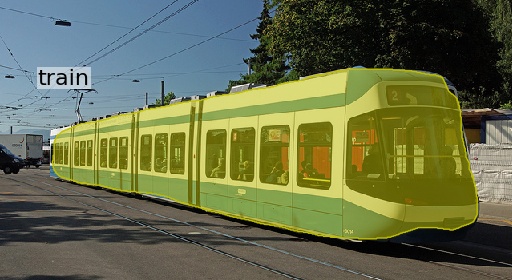}
\end{subfigure}%
\begin{subfigure}{.33\textwidth}
\centering
\includegraphics[width=\textwidth, height=4cm]{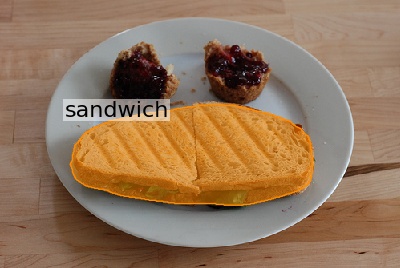}
\end{subfigure}
\caption{\textbf{Qualitative Visualizations.} Unseen category examples for the zero-shot segmentation task generated using our proposed ``$\text{M}{+}\text{CLIP}$'' variant (color $=$ object category).}
\label{fig:qualisegclip}
\end{figure*}

Please refer to Section \ref{sec:sota-comparison} of the main paper for our model definitions. We show unseen category qualitative visualizations for our model variants on the zero-shot detection task in Figures \ref{fig:qualidetw2v} and \ref{fig:qualidetclip}. Similarly, unseen category visualizations for our model variants on the zero-shot segmentation task are shown in Figures \ref{fig:qualisegw2v} and \ref{fig:qualisegclip}.

\newpage

\begin{table*}[t]
\setlength{\aboverulesep}{0pt}
\setlength{\belowrulesep}{0pt}
\setlength{\extrarowheight}{.75ex}
\setlength{\tabcolsep}{2.0pt}
\centering
\centering
\begin{tabular}{cc|ccccccccccccccccc|c}
\toprule
\toprule
\rowcolor{Gray}
\multicolumn{2}{c|}{Method} & \begin{tabular}[c]{@{}c@{}}air-\\plane\end{tabular} & bus & cat  & dog & cow & \begin{tabular}[c]{@{}c@{}}elep-\\hant\end{tabular} & \begin{tabular}[c]{@{}c@{}}umbr-\\ella\end{tabular} & tie & \begin{tabular}[c]{@{}c@{}}snow-\\board\end{tabular} & \begin{tabular}[c]{@{}c@{}}skate-\\board\end{tabular} & cup & knife & cake & couch & \begin{tabular}[c]{@{}c@{}}key-\\board\end{tabular} & sink & scissors & mAP  \\ \cline{1-20}

\multicolumn{1}{c|}{\multirow{3}{*}{\rotatebox[origin=c]{90}{F-RCNN}}} & RRFS \cite{huang2022robust} & 13.7 & 62.7 & 0.5 & 9.0 & \textbf{58.5} & 4.4 & 0.0 & \textbf{30.4} & 18.0 & 0.8 & 1.6 & 0.5 & 1.6 & \textbf{24.2} & 0.5 & 0.6 & 0.4 & 13.4 \\
\multicolumn{1}{c|}{} & Ours$_{\text{F}+\text{W2V}}$ & 19.1 & 53.2 & \textbf{43.3} & 20.6 & 22.7 & 24.2 & 0.4 & 0.0 & 29.0 & 2.0 & 10.4 & 2.8 & 6.3 & 9.2 & 2.7 & 4.5 & \textbf{2.0} & 14.8\\
\multicolumn{1}{c|}{}& Ours$_{\text{F}+\text{CLIP}}$ & \textbf{25.5} & \textbf{63.3} & 43.1 & \textbf{20.7} & 29.0 & \textbf{32.5} & \textbf{2.8} & 0.3 & \textbf{36.4} & \textbf{3.9} & \textbf{11.6} & \textbf{4.2} & \textbf{10.1} & 16.4 & \textbf{4.7} & \textbf{7.0} & 1.4 & \textbf{18.4}\\
\cdashline{1-20}
\multicolumn{1}{c|}{\multirow{3}{*}{\rotatebox[origin=c]{90}{M-RCNN}}}& Ours$_{\text{M}+\text{W2V}}$ & 20.3 & 51.7 & \textbf{46.8} & 19.3 & 21.5 & 24.6 & 0.9 & 0.0 & 26.4 & 2.4 & 10.2 & 2.6 & 7.8 & 13.5 & 3.9 & 5.4 & \textbf{3.0} & 15.3 \\
\multicolumn{1}{c|}{} & Ours$_{\text{M}+\text{CLIP}}$ & \textbf{21.7} & \textbf{63.7} & 39.7 & \textbf{21.6} & \textbf{27.1} & \textbf{31.8} & \textbf{2.7} & \textbf{0.3} & \textbf{33.7} & \textbf{3.4} & \textbf{10.8} & \textbf{5.5} & \textbf{11.0} & \textbf{21.7} & \textbf{7.3} & \textbf{6.3} & 0.7 & \textbf{18.2} \\
\multicolumn{1}{c|}{} &  &  &  &  &  &  &  &  &  &  &  &  &  &  &  &  &  &  &  \\
\bottomrule
\bottomrule
\end{tabular}
\caption{\textbf{Per category ZSD results for the $\mathbf{48/17}$ split.} We report the mAP for each novel category on the ZSD task for the $48/17$ split.}
\label{tab:perclass4817}
\bigskip

\begin{tabular}{cc|ccccccccccccccc|c}
\toprule
\toprule
\rowcolor{Gray}
\multicolumn{2}{c|}{Method} & airplane & train & \begin{tabular}[c]{@{}c@{}}parking\\ meter\end{tabular} & cat  & bear & \begin{tabular}[c]{@{}c@{}}suit-\\case\end{tabular} & frisbee & \begin{tabular}[c]{@{}c@{}}snow-\\board\end{tabular} & fork & sandwich & \begin{tabular}[c]{@{}c@{}}hot \\ dog\end{tabular} & toilet & mouse & toaster & \begin{tabular}[c]{@{}c@{}}hair\\ drier\end{tabular} & mAP  \\ \cline{1-18}
\multicolumn{1}{c|}{\multirow{5}{*}{\rotatebox[origin=c]{90}{F-RCNN}}}& PL \cite{rahman2018polarity}     & 20.0       & 48.2  & 0.6                                                     & 28.3 & 13.8 & 12.4     & 21.8    & 15.1      & 8.9  & 8.5      & 0.9                                                & 5.7    & 0.0   & \textbf{1.7}     & 0.0                                                  & 12.4 \\
\multicolumn{1}{c|}{}& SU \cite{hayat2020synthesizing}     & 10.1     & 48.7  & 1.2                                                     & 64.0 & \textbf{64.1} & 12.2     & 0.7     & 28.0      & 16.4 & 19.4     & 0.1                                                & 18.7   & 1.2   & 0.5     & 0.2                                                  & 19.0 \\
\multicolumn{1}{c|}{}& RRFS \cite{huang2022robust}   & 20.8     & \textbf{53.0}  & 1.3                                                     & 64.3 & {55.5} & 11.6     & 0.4     & 31.3      & 18.0 & 20.3     & 0.1                                                & 15.2   & \textbf{4.2}   & 0.5     & 0.6                                                  & 19.8 \\
\multicolumn{1}{c|}{}& Ours$_{\text{F}+\text{W2V}}$  &   20.0    &   38.8 &  5.7 &   \textbf{68.0}    &   19.6    &   20.0    &   \textbf{29.4}    &   \textbf{38.5}    & 11.3    &   17.8    &   0.6 &   20.8    &   1.9 &   1.0 &   0.2  & 19.6\\
\multicolumn{1}{c|}{}& Ours$_{\text{F}+\text{CLIP}}$  &   \textbf{29.7}    &   39.1 &  \textbf{18.3} &   62.3    &   52.0    &   \textbf{23.1}    &   21.0    &   38.3    & \textbf{20.5}    &   \textbf{20.6}    &   \textbf{13.6} &   \textbf{26.7}    &   0.8 &   0.2 &   \textbf{0.7}  & \textbf{24.5}\\
\cdashline{1-18}
\multicolumn{1}{c|}{\multirow{3}{*}{\rotatebox[origin=c]{90}{M-RCNN}}}& Ours$_{\text{M}+\text{W2V}}$ & 10.7 & 22.9 & 4.3 & \textbf{47.1} & 10.4 & 10.5 & \textbf{19.9} & \textbf{23.4} & 6.7 & 11.0 & 0.4 & 11.0 & \textbf{2.2} & \textbf{0.4} & 0.1 & 19.0 \\
\multicolumn{1}{c|}{}& Ours$_{\text{M}+\text{CLIP}}$  &   \textbf{18.2} & \textbf{26.3} & \textbf{11.6} & 42.2 & \textbf{45.9} & \textbf{13.3} & 18.7 & 22.3 & \textbf{13.3} & \textbf{14.0} & \textbf{8.1} & \textbf{17.4} & 0.4 & 0.2 & \textbf{1.4} & \textbf{25.1}\\
\multicolumn{1}{c|}{} &  &  &  &  &  &  &  &  &  &  &  &  &  &  &  &  &    \\
\bottomrule
\bottomrule
\end{tabular}
\caption{\textbf{Per category ZSD results for the $\mathbf{65/15}$ split.} We report the mAP for each novel category on the ZSD task for the $65/15$ split.}
\label{tab:perclass6515}
\end{table*}

\begin{table*}[t]
\setlength{\aboverulesep}{0pt}
\setlength{\belowrulesep}{0pt}
\setlength{\extrarowheight}{.75ex}
\setlength{\tabcolsep}{2.0pt}
\centering
\begin{tabular}{c|ccccccccccccccccc|c}
\toprule
\toprule
\rowcolor{Gray}
Method & \begin{tabular}[c]{@{}c@{}}air-\\plane\end{tabular} & bus & cat  & dog & cow & \begin{tabular}[c]{@{}c@{}}elep-\\hant\end{tabular} & \begin{tabular}[c]{@{}c@{}}umbr-\\ella\end{tabular} & tie & \begin{tabular}[c]{@{}c@{}}snow-\\board\end{tabular} & \begin{tabular}[c]{@{}c@{}}skate-\\board\end{tabular} & cup & knife & cake & couch & \begin{tabular}[c]{@{}c@{}}key-\\board\end{tabular} & sink & scissors & mAP  \\ \cline{1-19}
Ours$_{\text{M}+\text{W2V}}$ & 24.1 & 50.3 & \textbf{42.5} & 17.2 & 19.8 & 23.7 & 0.8 & 0.0 & 16.7 & 2.1 & 10.2 & 1.1 & 7.4 & 10.1 & 3.7 & 4.9 & \textbf{2.8} & 14.0\\
Ours$_{\text{M}+\text{CLIP}}$ & \textbf{27.8} & \textbf{64.1} & 36.6 & \textbf{19.7} & \textbf{24.5} & \textbf{30.8} & \textbf{2.8} & \textbf{0.2} & \textbf{21.7} & \textbf{2.9} & \textbf{10.5} & \textbf{3.5} & \textbf{10.8} & \textbf{20.0} & \textbf{8.2} & \textbf{6.3} & 0.2 & \textbf{17.1} \\
\bottomrule
\bottomrule
\end{tabular}
\caption{\textbf{Per category ZSI results for the $\mathbf{48/17}$ split.} We report the mAP for each novel category on the ZSI task for the $48/17$ split.}
\label{tab:perclassseg4817}
\bigskip

\begin{tabular}{c|ccccccccccccccc|c}
\toprule
\toprule
\rowcolor{Gray}
Method & airplane & train & \begin{tabular}[c]{@{}c@{}}parking\\ meter\end{tabular} & cat  & bear & \begin{tabular}[c]{@{}c@{}}suit-\\case\end{tabular} & frisbee & \begin{tabular}[c]{@{}c@{}}snow-\\board\end{tabular} & fork & sandwich & \begin{tabular}[c]{@{}c@{}}hot \\ dog\end{tabular} & toilet & mouse & toaster & \begin{tabular}[c]{@{}c@{}}hair\\ drier\end{tabular} & mAP  \\\cline{1-17}
Ours$_{\text{M}+\text{W2V}}$ & 13.0 & 25.3 & 5.0 & \textbf{48.3} & 12.2 & 10.2 & \textbf{21.4} & \textbf{14.8} & 2.5 & 10.7 & 0.4 & 11.2 & \textbf{2.3} & \textbf{0.4} & 0.1 & 17.9  \\
Ours$_{\text{M}+\text{CLIP}}$ & \textbf{17.7} & \textbf{26.8} & \textbf{12.9} & 42.5 & \textbf{43.9} & \textbf{13.4} & 19.8 & 14.3 & \textbf{6.5} & \textbf{12.7} & \textbf{6.7} & \textbf{17.9} & 0.4 & 0.2 & \textbf{1.3} & \textbf{23.5} \\
\bottomrule
\bottomrule
\end{tabular}
\caption{\textbf{Per category ZSI results for the $\mathbf{65/15}$ split.} We report the mAP for each novel category on the ZSI task for the $65/15$ split.}
\label{tab:perclassseg6515}
\end{table*}

\end{document}